\documentclass[letterpaper,twocolumn,10pt]{article}
\usepackage{usenix-2020-09}

\usepackage{tikz}
\usepackage{amsmath}

\usepackage{comment}
\usepackage{amssymb,amsfonts}
\usepackage{booktabs}
\usepackage{enumitem}
\usepackage{balance}
\usepackage{caption}

\usepackage{multirow}
\newcommand{\tabincell}[2]{\begin{tabular}{@{}#1@{}}\linespread{0.2}#2\end{tabular}}

\newcommand{\eg}{{e.g.}}
\newcommand{\ie}{{i.e.}}

\newcommand{\alibaba}{{Alibaba}}
\newcommand{\taobao}{{Mobile Taobao}}

\newcommand*{\email}[1]{\href{mailto:#1}{\nolinkurl{#1}}}

\begin{document}
%-------------------------------------------------------------------------------

%don't want date printed
\date{}

% make title bold and 14 pt font (Latex default is non-bold, 16 pt)

\title{\Large \bf Walle: An End-to-End, General-Purpose, and Large-Scale Production System for Device-Cloud Collaborative Machine Learning}

\author{
{\rm Chengfei Lv}\\
Zhejiang University \& Alibaba Group
\and
{\rm Chaoyue Niu\thanks{Chaoyue Niu is the corresponding author (\email {rvince@sjtu.edu.cn}).}}\\
Shanghai Jiao Tong University \& Alibaba Group
\and
{\rm Renjie Gu,\ Xiaotang Jiang, Zhaode Wang, Bin Liu, Ziqi Wu, Qiulin Yao, Congyu Huang,}\\ 
{\rm Panos Huang, Tao Huang, Hui Shu, Jinde Song, Bin Zou, Peng Lan, Guohuan Xu}\\
Alibaba Group
\and
{\rm Fei Wu}\\
Zhejiang University
\and
{\rm Shaojie Tang}\\
University of Texas at Dallas
\and
{\rm Fan Wu, Guihai Chen}\\
Shanghai Jiao Tong University
}

\maketitle

%-------------------------------------------------------------------------------
\begin{abstract}
%-------------------------------------------------------------------------------
To break the bottlenecks of mainstream cloud-based machine learning (ML) paradigm, we adopt device-cloud collaborative ML and build the first end-to-end and general-purpose system, called Walle, as the foundation. Walle consists of a deployment platform, distributing ML tasks to billion-scale devices in time; a data pipeline, efficiently preparing task input; and a compute container, providing a cross-platform and high-performance execution environment, while facilitating daily task iteration. Specifically, the compute container is based on Mobile Neural Network (MNN), a tensor compute engine along with the data processing and model execution libraries, which are exposed through a refined Python thread-level virtual machine (VM) to support diverse ML tasks and concurrent task execution. The core of MNN is the novel mechanisms of operator decomposition and semi-auto search, sharply reducing the workload in manually optimizing hundreds of operators for tens of hardware backends and further quickly identifying the best backend with runtime optimization for a computation graph. The data pipeline introduces an on-device stream processing framework to enable processing user behavior data at source. The deployment platform releases ML tasks with an efficient push-then-pull method and supports multi-granularity deployment policies. We evaluate Walle in practical e-commerce application scenarios to demonstrate its effectiveness, efficiency, and scalability. Extensive micro-benchmarks also highlight the superior performance of MNN and the Python thread-level VM. Walle has been in large-scale production use in Alibaba, while MNN has been open source with a broad impact in the community.
\end{abstract}

%-------------------------------------------------------------------------------
\vspace{-0.5em}
\section{Introduction}
\vspace{-0.2em}
%-------------------------------------------------------------------------------

To provide intelligent services for millions or even billions of smartphone users in industry, the mainstream paradigm lets mobile devices send requests with raw data and lets the cloud return results after data processing and model execution. However, this paradigm encounters three major bottlenecks: (1) {\em High Latency:} The network latency between each mobile device and the cloud plus the request processing latency of the cloud is in seconds, which is unacceptable for some real-time interactive applications. For example, the practical latency requirements of computer vision (CV), natural language processing (NLP), and recommendation tasks are in hundreds or even tens of milliseconds; (2) {\em High Cost and Heavy Load:} On the device side, uploading raw data will incur high cellular data usage, if Wi-Fi is not available. On the cloud side, receiving and storing enormous amounts of raw data from a massive number of mobile devices, processing data with diverse and sophisticated ML algorithms, and returning results in time, inevitably cause high overhead. For example, the size of 60s-long video or audio is in tens of MB, and the size of raw data for recommendation per user per day is in MB. Further multiplied by the scale of mobile devices, the total size of raw data is huge; and (3) {\em Data Security and Privacy:} Uploading the raw data with sensitive contents (\eg, personal information and user behaviors) may raise serious security and privacy concerns of users. Storing and processing raw data on the cloud may suffer from the risk of data breach.

By deconstructing the cloud-based ML paradigm, we can find that it simply regards mobile devices as interactive interfaces, but ignores the fact that mobile devices after 10 years of development can now undertake an appropriate load of data processing and model execution. Therefore, it does not leverage the natural device-side advantages of being close to users and data sources, thereby reducing latency and communication cost, mitigating the cloud-side load, and keeping private data on local devices. To overcome the bottlenecks of the mainstream cloud-based ML paradigm, the device-cloud collaborative ML paradigm emerged, which advocates offloading part of ML tasks to mobile devices and letting the cloud and the mobile devices collaboratively accomplish the ML tasks. Existing work tends to focus on the algorithmic decisions (\eg, device-cloud task splitting strategy \cite{proc:asplos17:dnn:split} and collaboration/interaction paradigm \cite{proc:aistats17:fl}) in either inference or training phase and normally for a specific application (\eg, video analytics \cite{proc:mlsys19:video:edge:classifier,proc:sigcomm20:video:diff,proc:sigcomm20:video:camera:resend}, text processing \cite{proc:mlsys19:fl:system}, recommendation \cite{proc:kdd21:patch,arxiv22:coda}). However, practical industrial scenarios tend to involve the full cycle of diverse CV, NLP, and recommendation applications to serve millions or even billions of mobile devices, building a general-purpose system to put device-cloud collaborative ML in large-scale production becomes a new requirement.

\begin{figure}[!t]
    \centering
    \includegraphics[width=.9\columnwidth]{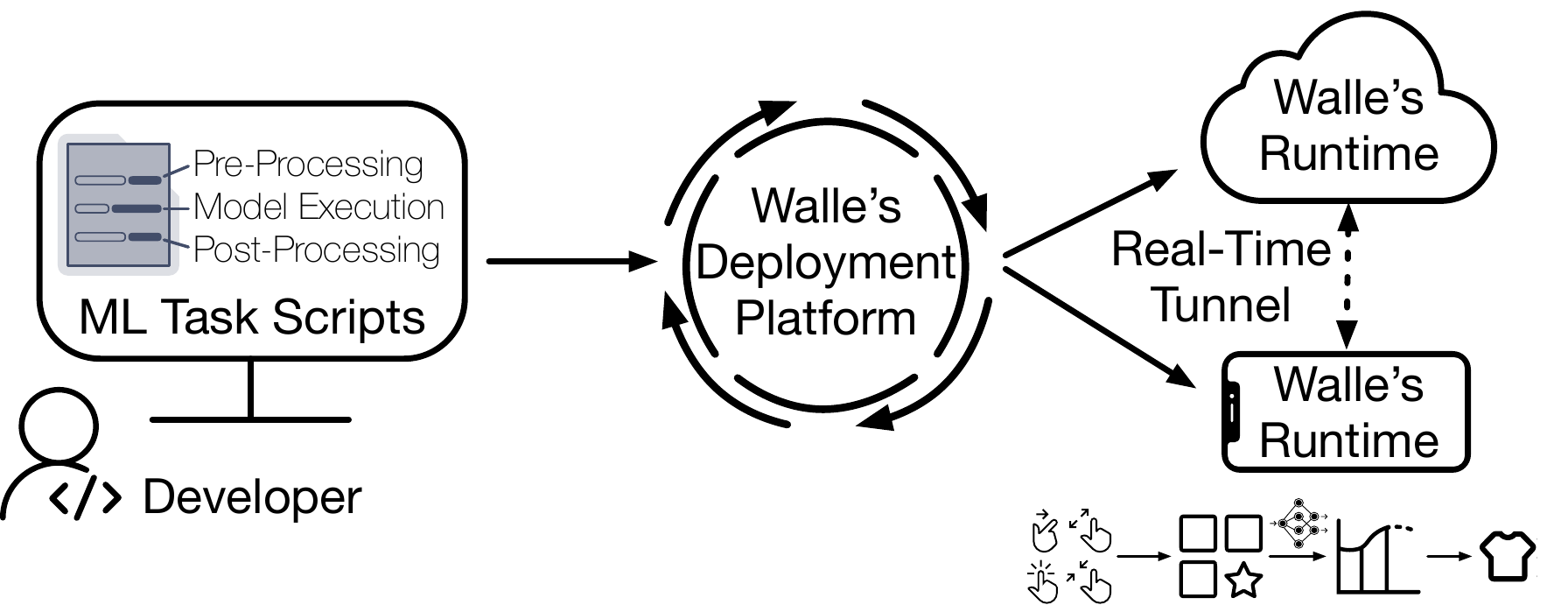}
    \vspace{-0.1em}
    \caption{Walle from the perspective of an ML task developer.} \label{fig:walle:workflow}
    \vspace{-1.3em}
\end{figure}

We build an end-to-end system, called Walle, the overall goal of which is to support general device-cloud collaboration (\eg, single device-cloud and multiple devices-cloud) in each phase of diverse ML tasks through exchanging any necessary content (\eg, data, feature, sample, model, model update, and intermediate result). As shown in Figure \ref{fig:walle:workflow}, Walle supports the whole cycle of ML tasks (\ie, pre-processing, model training and model inference, and post-processing) on both mobile devices and cloud servers in both development (\eg, the practical need of frequent experimentation and deployment for daily ML task iteration) and runtime (\ie, ML task execution and device-cloud data transfer) stages. By following the philosophy of building a general-purpose system rather than integrating massive application-specific or platform-specific solutions, Walle functions as a fundamental ML infrastructure with standard APIs and keeps the light-weight limit of mobile APPs, having supported 1,000+ kinds of CV, NLP, and recommendation tasks in large-scale production.

During building Walle, we encounter several practical requirements and challenges that motivate our design decisions. Walle is oriented by ML tasks and consists of a deployment platform, a data pipeline, and a compute container, catering to ML task deployment, input preparation, and execution, respectively. (1) For the compute container, one major requirement is to decouple ML task iteration from the monthly/weekly update of mobile APPs, making the classical method of integrating new functionalities into APPs infeasible. Another key requirement is to support diverse ML tasks with high performance across different operating systems (OS) and heterogeneous hardware of mobile devices and cloud servers. This requires to build a tensor compute engine in C/C++ and do operator-level and computation graph-level optimizations for each hardware backend. Two dominant strategies are manual optimization (\eg, in almost all ML engines), the workload of which is quite heavy that only some common cases can be covered; and auto tuning (\eg, in TVM \cite{proc:osdi18:chen}), which cannot support runtime optimization and is infeasible in industrial scenarios that involve massive heterogeneous devices or require frequent/quick ML task iteration. Based on the tensor compute engine, the libraries should be implemented to cover pre-processing, model training and inference, and post-processing as well as mobile devices and cloud servers in a unified way, rather than in a separate and incomplete way, like NumPy, OpenCV, TensorFlow (Lite), and PyTorch (Mobile). Without integrated design, the high performance of the tensor compute engine cannot be exposed to different libraries, the workload of optimizing each library on heterogeneous backends is heavy, and the package is large; (2) for the data pipeline, the overarching goal is to prepare raw data, which can come from different sources and are structured in various formats, as device-side or cloud-side ML model input. The mainstream paradigm of uploading all the device-side raw data to the cloud for aggregate processing is inefficient and error-prone; and (3) for the deployment platform, its key requirement is to manage, release, and deploy ML tasks for numerous mobile devices in a fine-grained, timely, and robust way, given massive ML task deployment requirements, intermittent device availability, and potential task failure.

We overcome the key challenges above and build Walle. (1) We choose dynamically-typed, widely-used Python as the script language for developing ML tasks in Walle and implement a Python VM as the core of the compute container by refining CPython in two aspects: one is to abandon the global interpreter lock (GIL) and support task-level multi-threading with VM isolation and data isolation; and the other is to perform tailoring for practical device-side need. Such design enables daily ML task iteration. At the bottom of the compute container, we implement a tensor compute engine along with standard data processing and model execution libraries, called MNN \cite{link:mnn}. MNN first introduces a novel geometric computing mechanism to decompose the transform and composite operators into atomic operators, thereby dramatically reducing the workload of manually optimizing hundreds of operators for tens of backends; and then introduces a novel semi-auto search mechanism to quickly identify the best backend with runtime optimization for a series of operators. At the top of the compute container, we expose MNN to Python thread-level VM as standard APIs, supporting the whole cycle of diverse ML tasks with standard data input. (2) For the data pipeline in Walle, we mainly build a new on-device stream processing framework to enable processing user behavior data at source. The key novelty is managing the trigger conditions of multiple stream processing tasks to generate different features with a trie structure for concurrent triggering. We also establish a real-time tunnel to transfer device-side fresh features to the cloud for use. (3) Regarding the deployment platform of Walle, we propose to manage task entity with git, categorize task-related files into shared and exclusive ones to facilitate multi-granularity deployments, and release tasks with an efficient push-then-pull method and in steps.

Walle is now part of Alibaba's ML backbone infrastructure, being invoked more than 153 billion times per day and supporting more than 0.3 billion daily active users, 30+ mobile APPs, and 300+ kinds of ML tasks. MNN is open source now with 6,600+ stars and 1,300+ forks on GitHub, and also is in production use in 10+ other companies. Evaluation of Walle in example real applications (\ie, livestreaming and recommendation) and platform statistics demonstrate effectiveness, efficiency, and scalability. Micro-benchmarks of MNN and Python thread-level VM show superiority.

We summarize the key contributions as follows: (1) Walle is the first end-to-end, general-purpose, and large-scale production system for device-cloud collaborative ML, masking hardware and software heterogeneity at the bottom, and supporting diverse ML tasks with daily iteration cycle and high performance at the top; (2) the compute container in Walle comprises MNN, which introduces geometric computing to sharply reduce the workload of manual operator-level optimization, and semi-auto search to identify the best backend with runtime optimization; and a Python VM, which is the first to abandon GIL and support task-level multi-threading, and also is the first to be ported to mobile devices; (3) the data pipeline in Walle introduces on-device stream processing with trie-based concurrent task triggering to enable processing user behavior data at source; and (4) the deployment platform in Walle supports fine-grained task release and deployment to billion-scale devices with strong timeliness and robustness.

%-------------------------------------------------------------------------------
\vspace{-0.2em}
\section{Preliminaries}
%-------------------------------------------------------------------------------

In this section, we first expound the background and motivation of building a general-purpose system for device-cloud collaborative ML. We then elaborate on the major design challenges. We finally draw the system requirements.  

\vspace{-0.3em}
\subsection{Background and Motivation}

{\bf Machine Learning Task.} From a developer's perspective, an ML task comprises scripts (\eg, codes in Python), resources (\eg, data, models, and dependent libraries), and configurations (\eg, trigger conditions mainly for specifying where and when to trigger the ML task). The whole workflow of an ML task can be divided into three phases or sub-tasks\footnote{We call ML sub-tasks also as ML tasks for convenience.}: pre-processing, model execution, and post-processing. In the pre-processing phase, raw data from multiple sources are cleaned, integrated, and processed to extract features and generate samples, which are then fed into models. In the model execution phase, a model is loaded to do training or inference. In the post-processing phase, the results of model inference are processed (\eg, by applying some ranking policies or business rules) to finally serve users.

{\bf Motivating Industrial Applications.} In Alibaba, there are now at least hundreds of online ML tasks to serve billion-scale daily active users with mobile devices in tens of business scenario, where CV, NLP, and recommendation tasks roughly account for 30\%, 10\%, and 60\% of the total tasks and run billion, one hundred billion, and billion times every day, respectively. In particular, (1) typical CV-type application scenarios include livestreaming, visual image search, short video analysis, augmented reality, and security checkup, where the major tasks include key frame detection, image segmentation and classification, item recognition, facial recognition and effects, human keypoint and pose detection, and porn detection; (2) typical NLP-type application scenarios include livestreaming and voice navigation, where the major tasks include automatic speech recognition, text to speech, text analysis, and text generation; and (3) typical recommendation-type application scenarios include item re-ranking, intelligent refresh, message popping, and page rearrangement, where the key tasks include click-through-rate prediction, click-conversion-rate prediction, user state recognition, and user intent detection.

\begin{comment}
\begin{figure}[!t]
    \centering
    \includegraphics[width=.9\columnwidth]{figures/Overview_V0.3.4.pdf}
    \caption{Device-cloud collaboration for ML tasks.}
\end{figure}
\end{comment}

{\bf Need for Device-Cloud Collaborative System.} The applications raise strict latency requirements on ML tasks. In general, (1) CV tasks need to process each image in 30ms; (2) NLP tasks require to process a 5s-long audio segment in 500ms or process an audio stream with latency less than 100ms; and (3) recommendation tasks need to generate outputs in 300ms. In addition, the raw data from massive users input to ML tasks are huge. For example, (1) for CV tasks, the size of a 60s-long, 1080p, and 8Mbps video is roughly 60MB; (2) for NLP tasks, the size of a 60s-long WAV/PCM audio is around 10MB; and (3) for recommendation tasks, one user normally  generates thousands of pieces of raw data per day, each piece in the size of KB. Furthermore, raw user data are more or less sensitive, raising security and privacy concerns.

The practical requirements above make the mainstream cloud-based ML paradigm infeasible and motivate us to adopt device-cloud collaborative ML. The key principle is that an ML task can be executed not only on the cloud but also on mobile devices, rather than purely on the cloud. During the execution of an ML task, mobile devices can work as a relay of the cloud, and vice versa. The choice of which side to execute which phase is flexible and should incorporate the practical need of the ML task and the characteristics of the cloud and mobile devices. For example, choosing which side for pre-processing should consider whether the side is near data source. Further observing the industrial need to support diverse ML tasks and massive devices, we are motivated to build an end-to-end and general-purpose system that can put device-cloud collaborative ML in large-scale production. 

\vspace{-0.5em}
\subsection{Practical Challenges}
\vspace{-0.2em}

A device-cloud collaborative ML system faces some practical challenges that span the execution, input preparation, and deployment stages of ML tasks as follows.  

{\bf Execution Challenges.} (1) {\em Long Iteration Cycle:} The common update cycle of a mobile APP includes development, testing, and integration of new functionalities (\eg, ML tasks to be deployed in our context), as well as APP store review and release to massive mobile devices in batches. As a result, most APPs is updated weekly, while some super APPs (\eg, \taobao, a shopping APP owned by Alibaba with 0.3 billion daily active users) are updated monthly. However, ML tasks require frequent experiments/deployments in nature, such that the effectiveness of different ML algorithms and models can be quickly verified, and the optimal feature combination and hyper-parameters can be identified; (2) {\em Heterogeneous Backends:} The cloud servers and mobile devices significantly differ in hardware (\eg, CPU, GPU, NPU, instruction set architecture (ISA), and memory) and OS (\eg, Android, iOS, Windows, and Linux). Among mobile devices, the ecosystem is even more fragmented; (3) {\em Diverse ML Tasks:} Industrial applications involve many kinds of ML tasks, requiring diverse model structures (\eg, convolutional neural network (CNN), recurrent neural network (RNN), transformer, generative adversarial network (GAN), and deep interest network (DIN)). Meanwhile, pre-processing and post-processing also involve lots of image, text, and numerical processing methods; and (4) {\em Limited Device Resources:} Each mobile APP has only one process. For \taobao, the maximum RAM is only 200MB, and the package size cannot exceed 300MB.

{\bf Input Preparation Challenges.} (1) {\em Atypical User Behavior Data:} For CV and NLP tasks, most raw data (\eg, image, video, text, and audio) are in standard formats, the pre-processing of which can be supported by standard libraries. Another major data source, the pre-processing of which cannot be directly supported, is each user's diverse behaviors in time and page series during interacting with a mobile APP and is essential to many ML (especially, recommendation) tasks. Conventionally, all the users' behavior data are uploaded to the cloud, far away from source, for stream processing with Flink. To enable pre-processing at source, there, however, does not exist an on-device stream processing framework; (2) {\em Diverse Trigger Conditions:} ML tasks tend to need many features. Each feature corresponds to a stream processing task and its trigger condition. How to efficiently manage multiple trigger conditions for concurrent task triggering is non-trivial.

{\bf Deployment Challenges.} (1) {\em Massive Task Deployment Requirements:} In Alibaba, the size of active ML tasks is at least in hundreds, and the mobile devices to be covered can reach the scale of billion. The release of each ML task also needs to incorporate APP versions, device-side and user-side differentiation; (2) {\em Intermittent Device Availability:} Mobile devices are with unstable wireless connections and allow only one APP to run on the foreground, while users tend to switch APPs frequently. Therefore, from the perspective of a certain APP, each device's availability is dynamic. Conventional push (\eg, based on persistent connection) or pull (\eg, based on polling) deployment method cannot guarantee timeliness and incurs high load on the cloud; (3) {\em Potential Task Failure:} A mobile APP runs as a single process. The failure of any task will lead to the crash of the whole APP, seriously impacting user experience. Further, due to the massive task deployment requirements, it is impractical to test each pre-release task on all relevant types of real devices.

\subsection{System Requirements}

Given the challenges above, the design of an device-cloud collaborative ML system should meet some requirements. 

The ML task execution environment needs to satisfy: (1) {\em Quick Task iteration:} ML tasks can be iterated daily on a mobile APP, reliving the dependence on the APP's original update cycle; (2) {\em Cross Platform:} OS-level and hardware-level heterogeneity should be masked; (3) {\em High Performance}: Optimization need to be specific to heterogeneous hardware backends of mobile devices and cloud servers; (4) {\em Universality:} Diverse CV, NLP, and recommendation tasks should be supported. The pre-processing, model execution, and post-processing phases of each ML task should be supported in an end-to-end way; and (5) {\em Light Weight:} The whole package size needs to be small, especially for mobile devices. 

The ML task input preparation pipeline needs to first introduce {\em a new on-device stream processing framework} with concurrent task triggering ability to enable processing user behavior data at source. To enable the cloud to consume the generated features (\eg, for feature fusion or model inference) far away from source with low latency, a real-time tunnel between mobile devices and the cloud also needs to be built. 

The ML task deployment platform should guarantee: (1) {\em Multi-Granularity:} Task release needs to support uniform, device-level grouping, user-level grouping, or even extremely device-specific policy; (2) {\em Timeliness:} A large number of mobile devices can be covered in short time; and (3) {\em Robustness:} Task deployment must put stability in the first place.

\begin{figure}[!t]
    \centering
    \includegraphics[width=0.97\columnwidth]{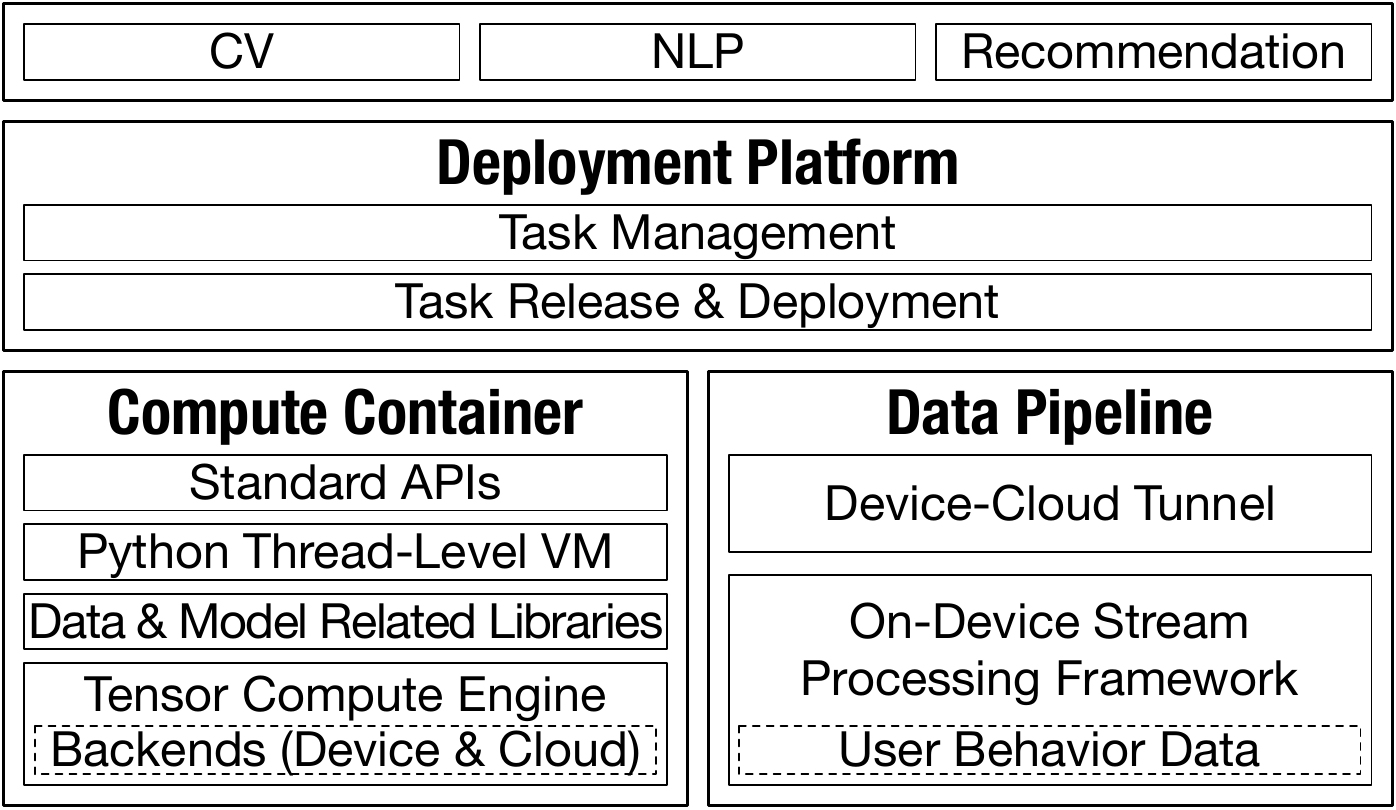}
    \vspace{-0.1em}
    \caption{Architecture of Walle.} \label{fig:walle_architecture}
    \vspace{-1em}
\end{figure}

\vspace{-0.2em}
\section{Walle: Architecture and Design Rationale}

Guided by the system requirements, we build Walle. We first introduce the whole architecture and then design rationale. 

\vspace{-0.2em}
\subsection{Architecture Overview}

As shown in Figure \ref{fig:walle_architecture}, the compute container in Walle comprises: (1) a cross-platform and high-performance tensor compute engine at the bottom; (2) data processing and model execution libraries based on the tensor compute engine; (3) a Python thread-level VM; and (4) standard APIs at the top. The data pipeline introduces: (1) an on-device stream processing framework; and (2) a real-time device-cloud tunnel. The deployment platform in Walle comprises: (1) a task management module; and (2) a task release and deployment module.

\vspace{-0.2em}
\subsection{Design Rationale}

\begin{figure}[!t]
    \centering
    \includegraphics[width=0.98\columnwidth]{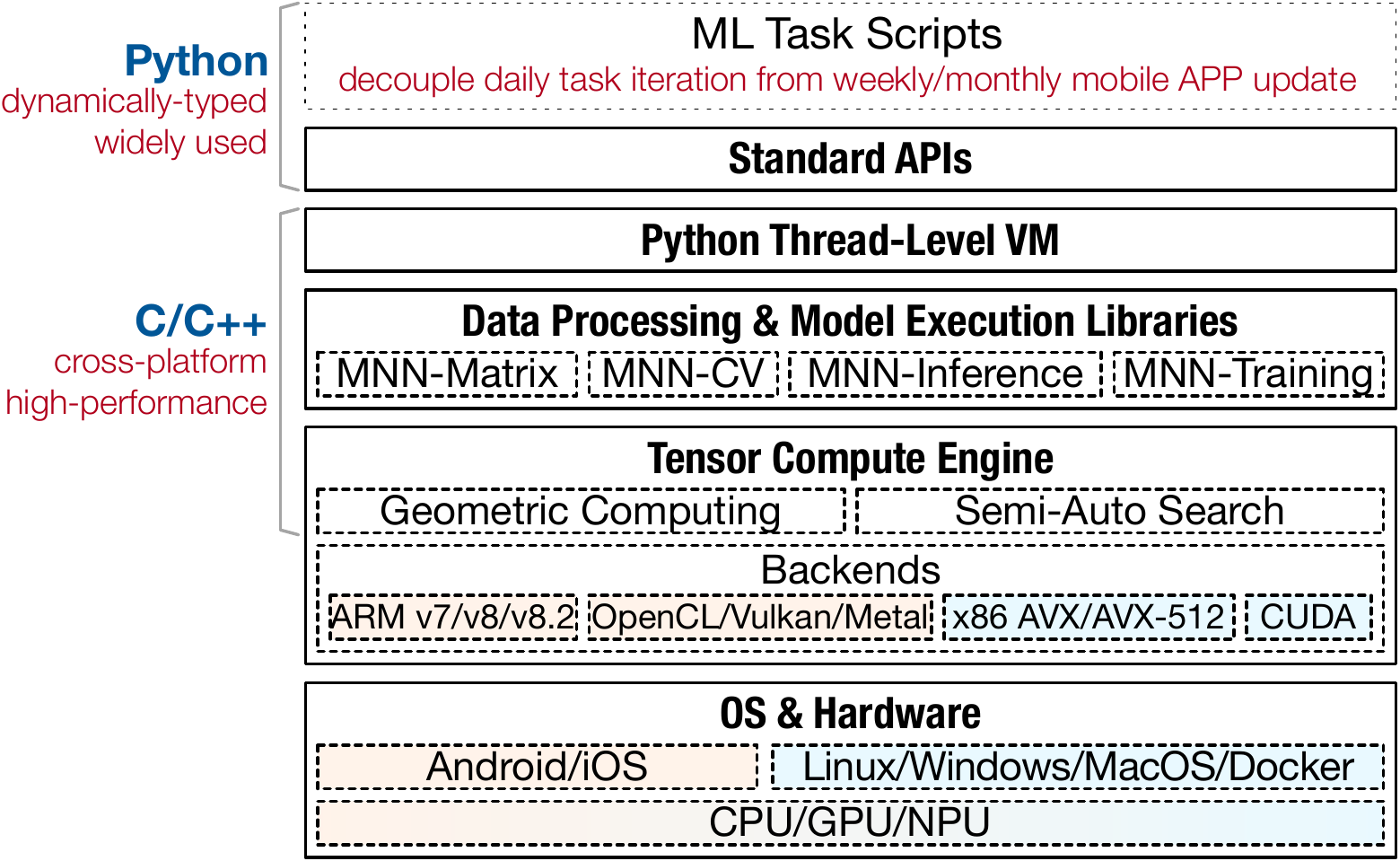}
    %\vspace{-0.1em}
    \caption{Architecture of compute container.} \label{fig:compute_container_architecture}
    \vspace{-1.2em}
\end{figure}

{\bf Rationale of Compute Container.} As shown in Figure \ref{fig:compute_container_architecture}, on the top, we choose Python as the script language, because Python is widely used in developing ML algorithms and also is a dynamically-typed and interpreted language. To support executing the Python scripts of ML tasks on different platforms, especially on resource-constraint mobile devices, we implement a Python VM by refining CPython and perform tailoring for the practical need of a mobile APP. Further considering the characteristics of ML task execution, including concurrent triggering of many tasks, independence across different tasks, and sequential execution of different phases in each individual ML task, we abandon GIL in Python VM and support task-level multi-threading by first binding each ML task with a thread and then conducting thread isolation. Such Python VM-based design endows the compute container with the capability of dynamic task delivery, decoupling daily ML task iteration from monthly/weekly mobile APP update. 

\begin{figure}[!t]
    \centering
    \includegraphics[width=0.96\columnwidth]{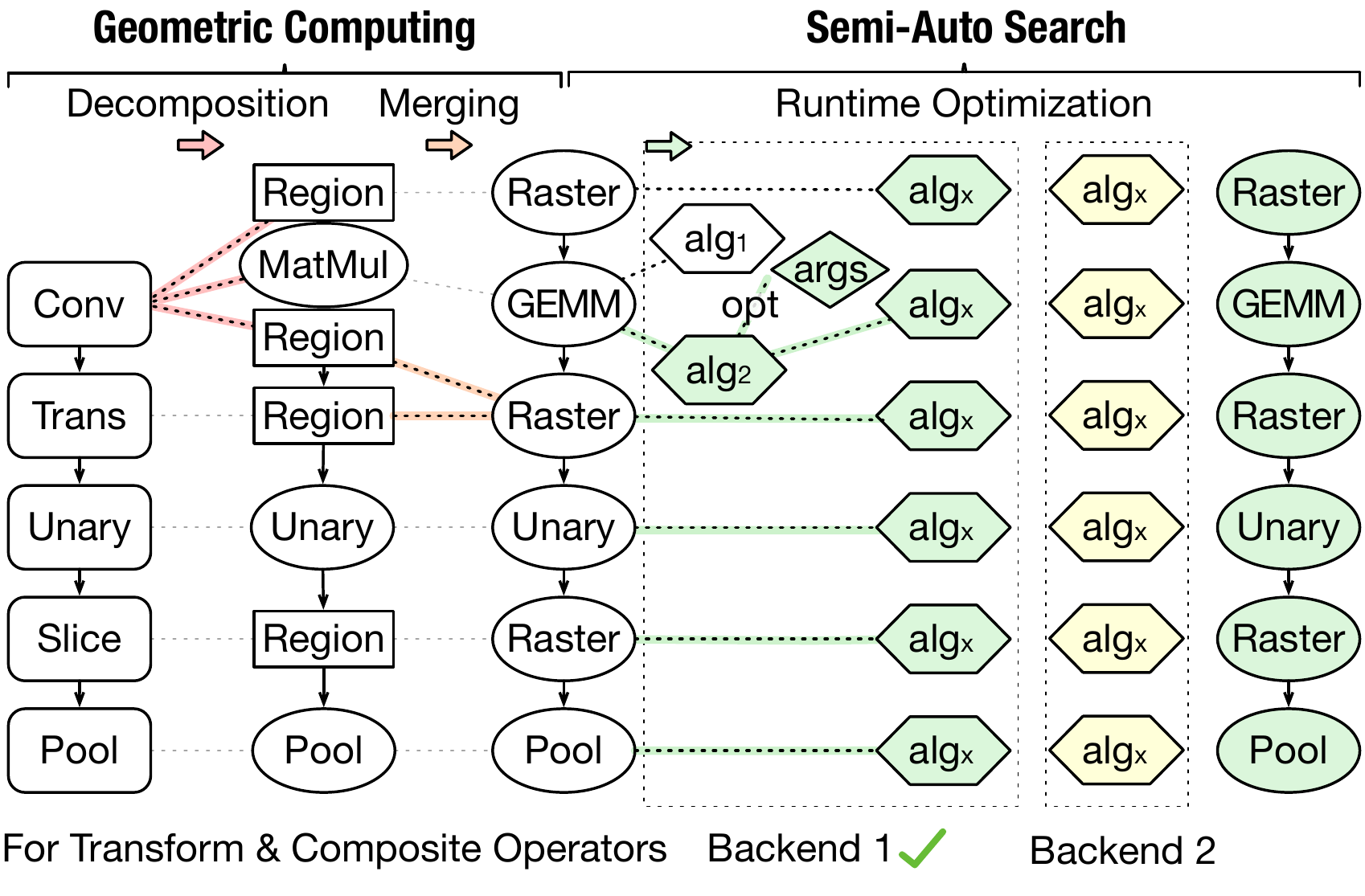}
    \vspace{-0.2em}
    \caption{Geometric computing and semi-auto search.}
    \vspace{-1.2em}
    \label{fig:geo_compute_search}
\end{figure}

At the bottom, we implement a tensor compute engine in C/C++ for cross-platform and high-performance considerations. The cores are the novel mechanisms of geometric computing and semi-auto search, as shown in Figure~\ref{fig:geo_compute_search}. In particular, geometric computing extracts a new atomic operator from transform operators, by leveraging the nature of coordinate transformation as well as the linear mapping between an element's coordinate and its memory address. As a result, all the transform and composite operators, accounting for roughly 49\% of all the operators, can be decomposed to the atomic operators, reducing 46\% of the workload of manually implementing and optimizing 124 operators for 16 kinds of backends from algorithm, ISA, memory, and assembly. Then, to quickly identify the backend available on a mobile device or a cloud server to execute a computation graph with a series of operators at the minimum cost, semi-auto search is applied in runtime to find the optimal implementation algorithm with the optimal parameters for each operator on each available backend. The parameter search is converted to solving a constrained optimization problem, by incorporating the hardware properties of the backend and the sizes of the implementation algorithm's inputs. Based on the tensor compute engine, we implement the libraries of scientific computing, image processing, model inference, and model training, and expose them to Python VM as standard APIs, supporting the whole cycle of diverse ML tasks with standard data input.

\begin{figure}[!t]
    \centering
    \includegraphics[width=0.95\columnwidth]{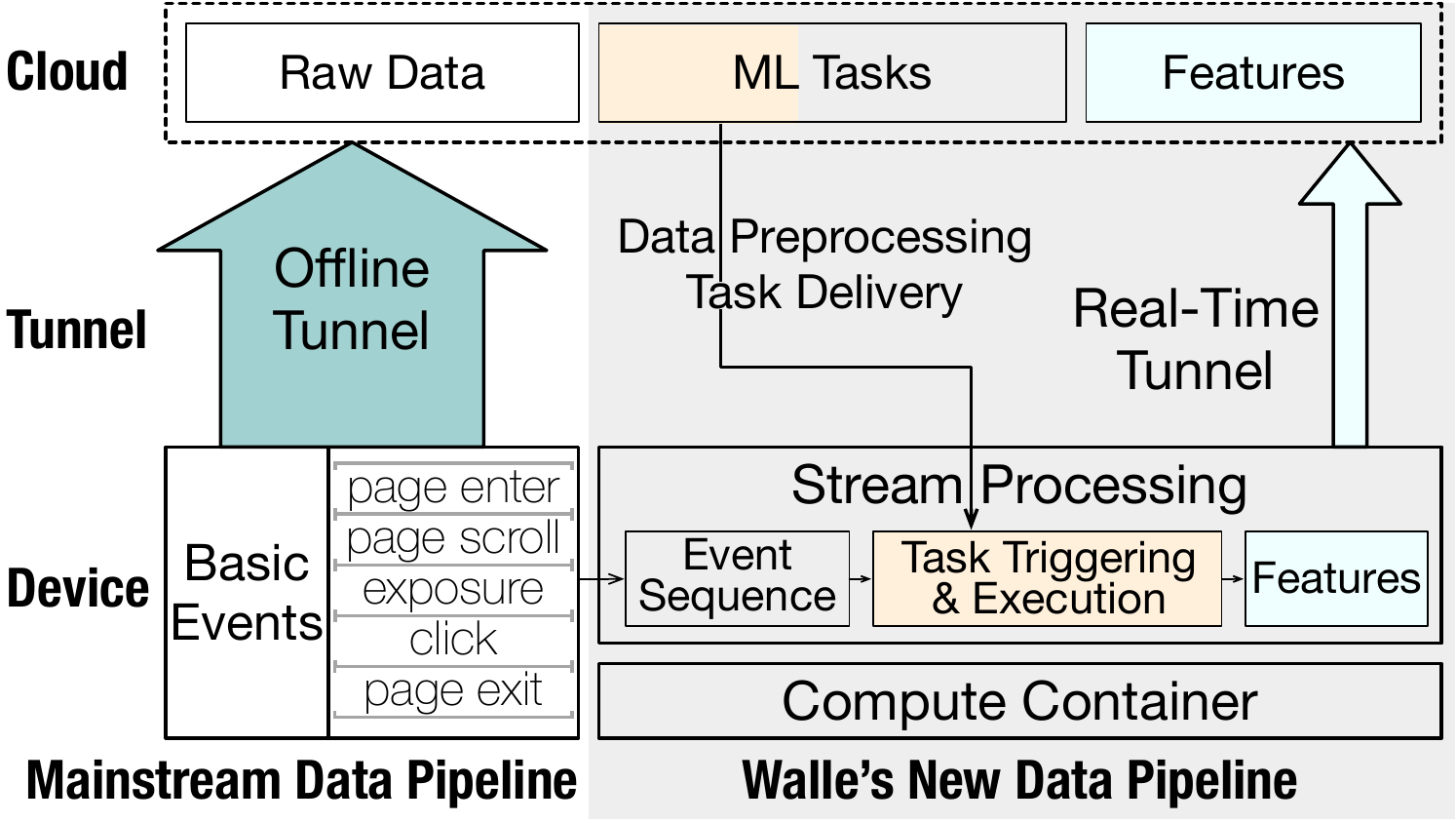}
    \vspace{-0.1em}
    \caption{Architecture of data pipeline.}\label{fig:data_pipeline_arch}
    \vspace{-1.5em}
\end{figure}

{\bf Rationale of Data Pipeline.} The architecture is depicted in Figure \ref{fig:data_pipeline_arch}. First, a user's behaviors are naturally recorded as a time-level event sequence, based on which the page-level event sequence can be created by aggregating the events in the same pages. Then, the trigger condition of a stream processing task can be specified by a sequence of event/page ids. To support concurrent triggering, we model matching multiple trigger conditions with the event sequence as a string matching problem with multiple wildcard patterns and propose to organize trigger conditions with a trie, such that if a new event comes, all the triggered tasks can be picked out for execution. Given a stream processing task can be triggered frequently over the continuously generated event sequence, while the size of one-time output is small, we design a collective storage mechanism to reduce the frequency of write. Finally, to upload the output of on-device stream processing with low latency, we leverage persistent connection to implement a real-time tunnel, transferring up to 30KB data within 500ms.

{\bf Rationale of Deployment Platform.} We first manage the task entity with git and categorize task-related files into shared and exclusive ones, according to how many devices can use the files in common. The file categorization further facilitates the uniform and customized policies of task deployment. To guarantee the timeliness of task deployment, we propose a novel push-then-pull method based on transient connection, where the push functionality reuses the existing client-side http request for business services, while the pull functionality is via content delivery network (CDN) and Alibaba cloud enterprise network (CEN). For the robustness of task deployment, we introduce task simulation test with the cloud-side compute container before release and enforce releasing task in steps, while allowing rollback in the case of task failure. 

In what follows, we present the design and implementation details of the compute container in Section \ref{sec:walle:compute:container}, the data pipeline in Section \ref{sec:walle:data:pipeline}, and the deployment platform in Section \ref{sec:walle:deployment:platform}.

\vspace{-0.2em}
\section{Compute Container in Walle }\label{sec:walle:compute:container}
\vspace{-0.1em}

We introduce the compute container in a bottom-up way: MNN, a tensor compute engine along with the data processing and model execution libraries; Python thread-level VM; and standard APIs of MNN.

\vspace{-0.3em}
\subsection{Tensor Compute Engine}\label{subsec:tensor:compute:engine}

Tensor computation can be viewed as the basis of data processing and ML, and the operators of underlying tensor computation can be divided into four categories: (1) {\em Atomic Operators}, which function as the basic unit of backend optimization, such as some common unary operators (\eg, taking square) and binary operators (\eg, addition, subtraction, multiplication, and division); (2) {\em Transform Operators}, which change the shape and/or reorder the elements, such as transpose, slicing, concatenation, and permutation; (3) {\em Composite Operators}, which can be decomposed into the atomic and transform operators, such as 3D convolution and pooling, normalization, exponential linear unit, and long short-term memory cell; and (4) {\em Control-Flow Operators}, including if and while.

{\bf Geometric Computing.} Currently, MNN can support $N_{aop} = 61$ atomic operators, $N_{top} = 45$ transform operators, $N_{cop} = 16$ composite operators, and $N_{fop} = 2$ control-flow operators. The workload of implementing and optimizing the operators for all $N_{ba} = 16$ backends in MNN is $O((N_{aop} + N_{top} + N_{cop}) \times N_{ba} + N_{fop} = 1954)$. Further considering the workload involving the atomic and control-flow operators is unavoidable, we turn to reducing the workload involving the transform and composite operators, which roughly accounts for half of the whole load and will grow in the future (\eg, as more composite operators are required to support more kinds of deep neural network (DNN)). Our key idea is to extract a new atomic operator, called ``raster'', from the transform operators. Then, both the transform operators and the composite operators can be decomposed into the raster operator and the atomic operators. Since only the atomic and raster operators need to be optimized for each backend, the whole workload becomes $O((N_{aop} + 1) \times N_{ba} + N_{top} + N_{cop} + N_{fop} = 1055)$, reducing roughly $46\%$ of the workload. Now, the problems become what is the raster operator and how to implement it. We propose a geometric computing mechanism as follows.  

In essence, the basic functionality of the transform operators is to move an element from a memory address to another memory address, or from geometry, is to transform the coordinate of the element to another coordinate. In addition, the memory address is a deterministic linear function of the coordinate. Moreover, given a certain transform operator, the formula of coordinate transformation can be determined. As a result, with the coordinate of an element in the input or output tensor, typically the element's index in the input or output tensor, the original memory address and the memory address after movement can also be determined. The raster operator is introduced to move the elements between the input and output tensors according to the memory addresses and by traversing the coordinates. We take slicing for example. $A$ is a $2 \times 4$ matrix, placed in contiguous memory addresses with a unique identifier/pointer. The slicing of $A$ by leaving only the second row is denoted as $B$, which is a $1 \times 4$ matrix. For an element $B_{i,j}$ with the row index $i$ and the column index $j$ (\ie, the coordinate $(i, j)$) in $B$, its memory identifier relative to $B$'s unique identifier is $i \times 4 + j$, which is linear with the coordinate, where the coefficients $(4, 1)$ are called the strides. According to the definition/rule of slicing (\ie, $B_{i,j} = A_{i + 1, j}$), the coordinate of the corresponding element $A_{i + 1, j}$ in $A$ is $(i + 1, j)$, and the relative memory identifier is $(i + 1) \times 4 + j = 4i + j + 4$, where the coefficients $(4, 1)$ are the strides, and the intercept 4 is called offset. The raster operator can realize the functionality of slicing by iterating the coordinates $\{(i, j)| 0 \leq i < 1, 0 \leq j < 4, i, j \in \mathbb{Z}\}$ and moving each $A_{i+1, j}$ to $B_{i,j}$ using their memory addresses.

In practical implementation of the raster operator, we introduce a supporting concept, called ``region'', which contains an input tensor, the range of coordinate, as well as the linear mappings between an element's coordinate and its memory addresses in the input and output tensors, which are called ``views'' and can be specified by the strides and offsets. In addition, after operator decomposition, some raster operations can be merged for optimization. One policy is called vertical merging, which mainly deals with two successive raster operations, skips indirect references, and operates on the original tensor; and the other policy is called horizontal merging, which handles two parallel raster operations with the same region and keeps only one raster operation. 

{\bf Atomic Operator Optimization.} Specific to the atomic operators, including the raster operator, we incorporate hardware heterogeneity and optimize the implementation from the perspectives of algorithm, ISA, memory, and assembly. (1) The algorithm-level optimization is specific to some compute-intensive operators, typically convolution and matrix multiplication. We take more efficient algorithms, including Winograd and Strassen algorithms, to sharply reduce the number of multiplications; (2) the ISA-level optimization leverages single instruction multiple data (SIMD), such as ARM Neon and x86 AVX512, for speedup. To adequately exploit data-level parallelism in SIMD, we carefully design data layout and data packing. Specifically, we take a new NC/4HW4 layout \cite{link:data:layout} and a channel-major packing for convolution; (3) the memory-level optimization focuses mainly on reducing the number of read and write as well as improving the contiguity of memory allocation. In particular, for matrix multiplication, we apply tiling and memory reordering; and (4) the assembly-based optimization can achieve instruction-level speedup. We implement core operators with hand-written assembly codes and carefully apply some optimizations, such as loop unrolling, software pipelining, and instruction reordering.

{\bf Semi-Auto Search.} Data processing and model execution normally involve a series of operators (\ie, the atomic, raster, and control-flow operators after decomposition). Meanwhile, different backends have different implementations and optimizations for the operators, and a mobile device or a cloud server tends to have several backends available. The global goal of semi-auto search is to identify the backend with the minimum cost. The cost of each backend is the sum of all the operators with the optimal implementations. To identify the optimal implementation algorithm for a certain operator on a certain backend, the optimal parameters of each possible algorithm need to be found. This is converted to a constrained optimization problem that can be quickly solved, where the objective is computation or memory cost, and the constraints incorporate the hardware constraints of the backend and the sizes of the algorithm's inputs. We formulate the whole process of semi-auto search and introduce the details as follows.

We let $BA$ denote the set of all available backends and let ${op}_1 \rightarrow {op}_2 \rightarrow \ldots \rightarrow {op}_n$ denote the series of $n$ operators for execution. The cost of a backend $ba \in BA$ is defined as
\begin{align}
    C_{ba} = \sum_{i=1}^{n} C_{op_i, ba},
\end{align}
where $C_{op_i, ba}$ denotes the cost of the operator $op_i$ with the optimal implementation on the backend $ba$. The goal of semi-auto search is to find the backend with the minimum cost, which can be expressed as
\begin{align}
    {\arg\min}_{ba \in BA} C_{ba}. 
\end{align}
Then, the problem is how to compute each $C_{op_i, ba}$. For each operator $op_i$ and the backend $ba$, we let $algs(op_i, ba)$ denote all feasible implementation algorithms with the optimal parameters. Then, $C_{op_i, ba}$ is defined as
\begin{align}
    C_{op_i, ba} = \min_{alg \in algs(op_i, ba)} \frac{Q_{alg}}{P_{ba}} + S_{alg, ba},
\end{align}
where (1) $Q_{alg}$ denotes the number of elementary calculations in the algorithm $alg$, which can be obtained given the (``optimal'' here) parameters and the sizes of the inputs; (2) $P_{ba}$ represents the performance of the backend $Ba$. In MNN, for a CPU-type backend, if the backend $ba$ supports ARMv8.2-FP16, $P_{ba}$ empirically takes 16 times the frequency; otherwise, $P_{ba}$ takes 8 times the frequency. For a GPU-type backend, $P_{ba}$ is empirically set to the number of floating point operations per second (FLOPS) by manual testing; and (3) $S_{alg, ba}$ denotes the scheduling cost of the algorithm $alg$ on the backend $ba$. In MNN, for a CPU-type backend, $S_{alg, ba}$ is set to $0$; and for a GPU-type backend, $S_{alg, ba}$ is empirically set and mainly considers the time of data transfer. Now, the remaining problem is for an operator ${op}_i$, a backend $ba$, an implement algorithm $alg$, and the sizes of the inputs, how to determine the optimal parameters of the algorithm. In practice, we formulate it into a constrained optimization problem, where the objective is to minimize the computation or memory cost, and the constraints mainly include the width of SIMD unit, the number of registers, the number of threads, and the sizes of the inputs. In addition, we focus mainly on optimizing the following parameters: the packing size in SIMD, the tile size in matrix multiplication, the block unit in the Winograd algorithm, and the reduction of the elementary calculations using the Strassen algorithm. We take the optimization of the title size in matrix multiplication for example. We let $A$ denote an $a \times e$ matrix, let $B$ denote an $e \times b$ matrix, let $t_e$ denote the tile size along the axis with the equal size, let $t_b$ denote the tile size along the axis of $B$'s columns, and let $N_r$ denote the number of registers. The optimization objective is minimizing the times of memory read and write. The formula of the optimization problem is given as follows:
\begin{align}
\begin{aligned}
    \min_{t_e, t_b} \quad &\frac{e}{t_e} \times \frac{b}{t_b} \times \left(a \times t_e + a \times t_b +  t_e \times t_b \right),\\
    \text{s.t.} \quad &t_e \times t_b + t_e + t_b \leq N_r,  
\end{aligned}
\end{align}
which can be solved efficiently in runtime. 

Compared with manual search, which optimizes the implementation algorithms with some common parameters for each operator case by case, semi-auto search not only can sharply reduce the workload but also can find the optimal parameters with higher probabilities. Regarding why not adopt auto tuning in TVM, it does not exploit manual experience in operator optimization, consumes long time of static compilation due to the large search space at the operator and graph levels for a certain backend, and cannot support runtime optimization. Most importantly, given the restriction on executable files and just-in-time (JIT) compilation on iOS devices for security \cite{link:ios:develop}, the compiled models generated by TVM must be linked into mobile APPs with monthly/weekly update and cannot be daily iterated as desired. Therefore, TVM is infeasible in industrial applications that involve a large number of heterogeneous devices or require frequent/quick task iteration (\eg, updating the deployed ML models). In contrast, our design of the tensor compute engine essentially leverages manual operator-level optimization for heterogeneous backends to narrow down the space of semi-auto search, thereby supporting deploying models as regular resource files and further facilitating runtime optimization and daily ML task iteration in Python VM. Another benefit is that the package size of mobile APPs will not increase in the long term for more and more ML tasks.

\vspace{-0.7em}
\subsection{Data and Model Related Libraries}
\vspace{-0.1em}

With the tensor compute engine, we implement the libraries of scientific computing and image processing for the pre-processing and post-processing phases of an ML task, as well as the libraries of model inference and model training. In particular, the scientific computing and image processing libraries can be regarded as the optimized implementations of NumPy \cite{jour:nature20:numpy} and OpenCV \cite{link:opencv} in terms of light weight and high performance. The light weight means that the sizes of libraries can be reduced without manual tailoring. The original sizes of NumPy 1.9.3 and OpenCV 3.4.3 are 2.1MB and 1.2MB, and decrease to 51KB and 129KB in MNN, respectively. The high performance indicates that the performance optimization of the underlying tensor compute engine can be inherited to the libraries, avoiding the extra workload. We introduce the implementations of the libraries as follows.  

{\bf Scientific Computing \& Image Processing.} We use the atomic, raster, and control-flow operators to support array creation and manipulation routines, binary operations, linear algebra, logic functions, padding arrays, random sampling, mathematical functions, etc, in the scientific computing library; and to support image filtering, geometric and miscellaneous image transformations, drawing functions, color space conversions, etc, in the image processing library. 

{\bf Model Inference \& Model Training.} We currently provide two modes of model inference in MNN, called session and module. The module mode can support the control-flow operators, which are required by transformer, dynamic RNN, etc, whereas the session mode cannot. The session-based model inference can be divided into four steps: (1) load a model, create a session, arrange all the operators in the computation graph according to the topological ordering, and apply for the tensors that all the operators need; (2) given the shape of each input tensor and the definition of each operator, compute the shapes of all the tensors; (3) perform geometric computing, particularly, first decompose the transform and composite operators into the atomic and raster operators, and then do vertical and horizontal merging for raster operators; and (4) identify the optimal backend with semi-auto search, request memory for each operator and execute in sequence, and return the inference result. In the second step, the control-flow operators require the intermediate result to determine the following execution order and thus cannot be supported in the session mode. To solve this problem, when loading the model in the first step, the module mode splits the computation graph into modules (\ie, sub-graphs) iteratively, according to the positions of the control-flow operators. Then, the execution of each module is the same as that of the session. 

We implement model training by adding two common optimizers: stochastic gradient descent (SGD) and adaptive moment estimation (ADAM). At the bottom, we add the gradient operators of all the atomic operators and one raster operator.

\begin{figure}[!t]
    \centering
    \includegraphics[width=0.95\columnwidth]{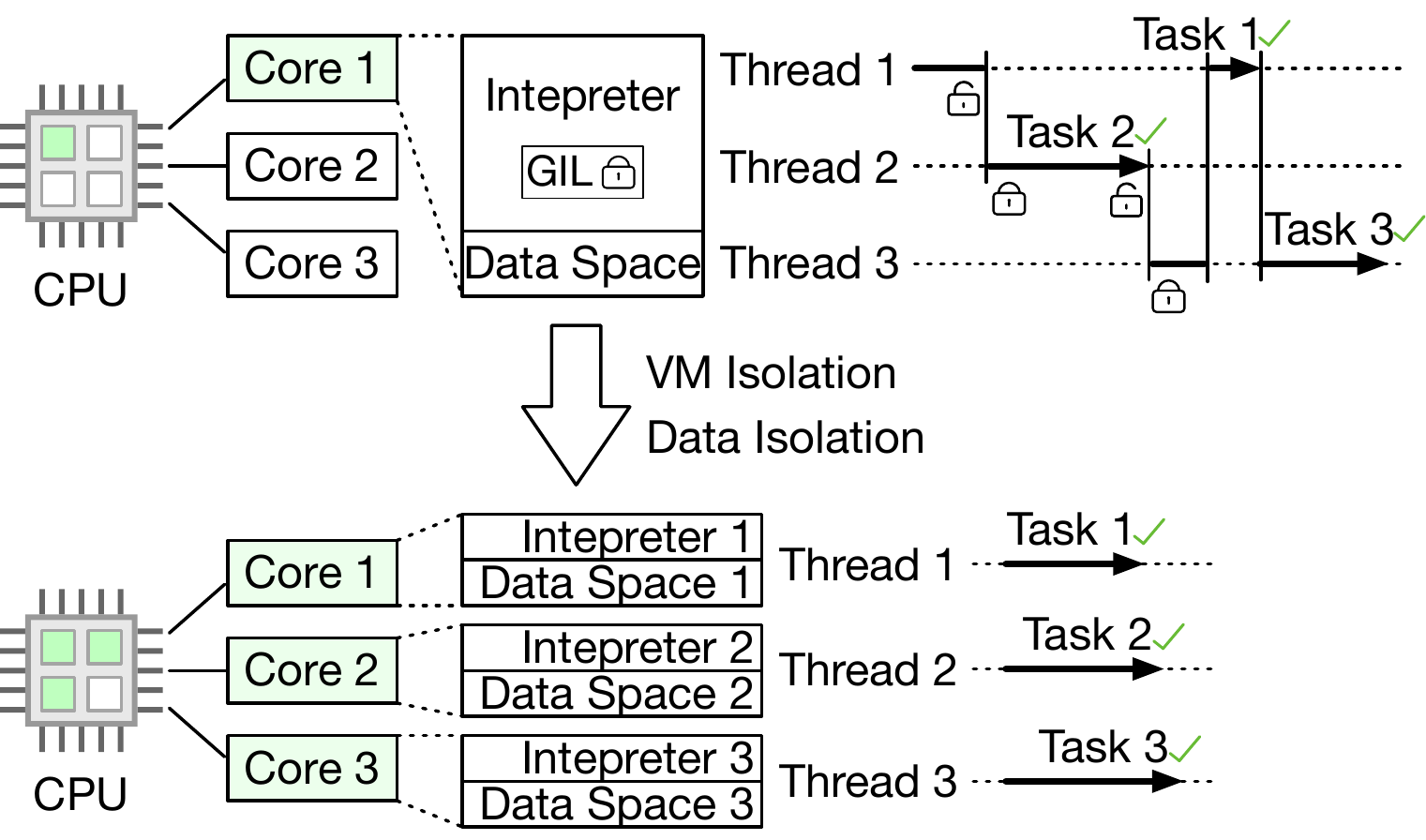}
    \vspace{-0.1em}
    \caption{Python thread-level virtual machine.}\label{fig:pythonvm}
    \vspace{-1.1em}
\end{figure}

\vspace{-0.3em}
\subsection{Python Thread-Level Virtual Machine}

Most ML tasks are implemented in Python and require a Python VM to execute the Python scripts. We choose the official and the most widely-used Python compiler and interpreter, called CPython \cite{link:cpython}. However, there exist two key problems in the porting process of CPython, especially for resource-constraint mobile devices. The first problem is that the size of the package is large. For example, CPython 2.7.15 contains 500+ scripts in C and 1,600+ libraries, including many redundant functionalities for mobile APPs. The second problem is that CPython cannot support multi-threading to improve efficiency. CPython originally cannot support concurrent programming and introduces GIL for multi-processing. GIL allows only one thread to be processed at one time within a process. However, each mobile APP has only one process and does not allow multi-processing. How to support task-level multi-threading in Python VM becomes a problem.

To reduce the package size, we tailor the functionalities, libraries, and modules for the practical need of \taobao. (1) {\em Functionality Tailoring:} CPython first compiles Python code into bytecode with the file suffix ``.pyc'' and then interprets the bytecode for execution. By leaving the compile phase on the cloud and sending only the bytecode to mobile devices for execution, we can delete all the compile modules, saving 17 scripts in C. (2) {\em Library and Module Tailoring:} We keep 36 necessary libraries (\eg, abc, type, re, and functools) and 32 modules (\eg, zipimport, sys, exceptions, and gc). After package tailoring, we implement a light-weight Python interpreter for mobile devices, which is the first in industry. For example, on ARM64-based iOS, the package size decreases from 10MB+ to only 1.3MB.

Regarding multi-threading, we abandon GIL and further design and implement the first Python thread-level interpreter in industry, supporting the concurrent execution of many tasks. As shown in Figure \ref{fig:pythonvm}, each task is scheduled to a certain thread, which creates an independent VM and contains the VM runtime and task-related data. For thread safety, the key is to perform thread-level VM isolation and data isolation, which pin a VM to its thread and further pin the context of VM runtime to the thread. (1) {\em VM Isolation:} The lifecycle of the original Python VM is pinned to the process, each process having one VM. We need to modify the creation of VM instances such that a process can hold multiple thread-level VMs, each VM having its independent lifecycle. In CPython, VM is defined as a struct in C, called PyInterpreterState. When CPython starts, one PyInterpreterState instance will be initialized. We modify the initialization of CPython, particularly creating and initializing a PyInterpreterState instance for each thread. (2) {\em Data Isolation:} Besides VM itself, the context of VM runtime (\eg, type system, module, and task-related data) should also be isolated on the level of thread, avoiding the concurrency problem of multi-threading without the protection of GIL. We adopt the thread-specific data (TSD) technique for data isolation, such that each thread has its own data space, and different threads cannot access the same data simultaneously. We mainly apply TSD to type system, buffer pool, object allocation, and garbage collection. 
\vspace{-0.4em}
\subsection{Standard APIs}

We expose the cross-platform libraries of data processing and model execution through Python VM to support ML tasks. For pre-processing and post-processing, the scientific computing and image processing APIs are consistent with the original APIs of NumPy and OpenCV to be developer-friendly, such as matmul, swapaxes, concatenate, split, resize, warpAffine, warpPerspective, cvtColor, GaussianBlur, etc. For model inference and model training, the APIs of common model-level and data-level operations are exposed, such as data loading, model loading and saving, session creation and execution, optimizers, hyper-parameter setting, loss computing, etc.

\vspace{-0.2em}
\section{Data Pipeline in Walle}\label{sec:walle:data:pipeline}

We detail the on-device stream processing framework and the real-time device-cloud tunnel in the data pipeline. 

\begin{figure}[!t]
    \centering
    \includegraphics[width=0.95\columnwidth]{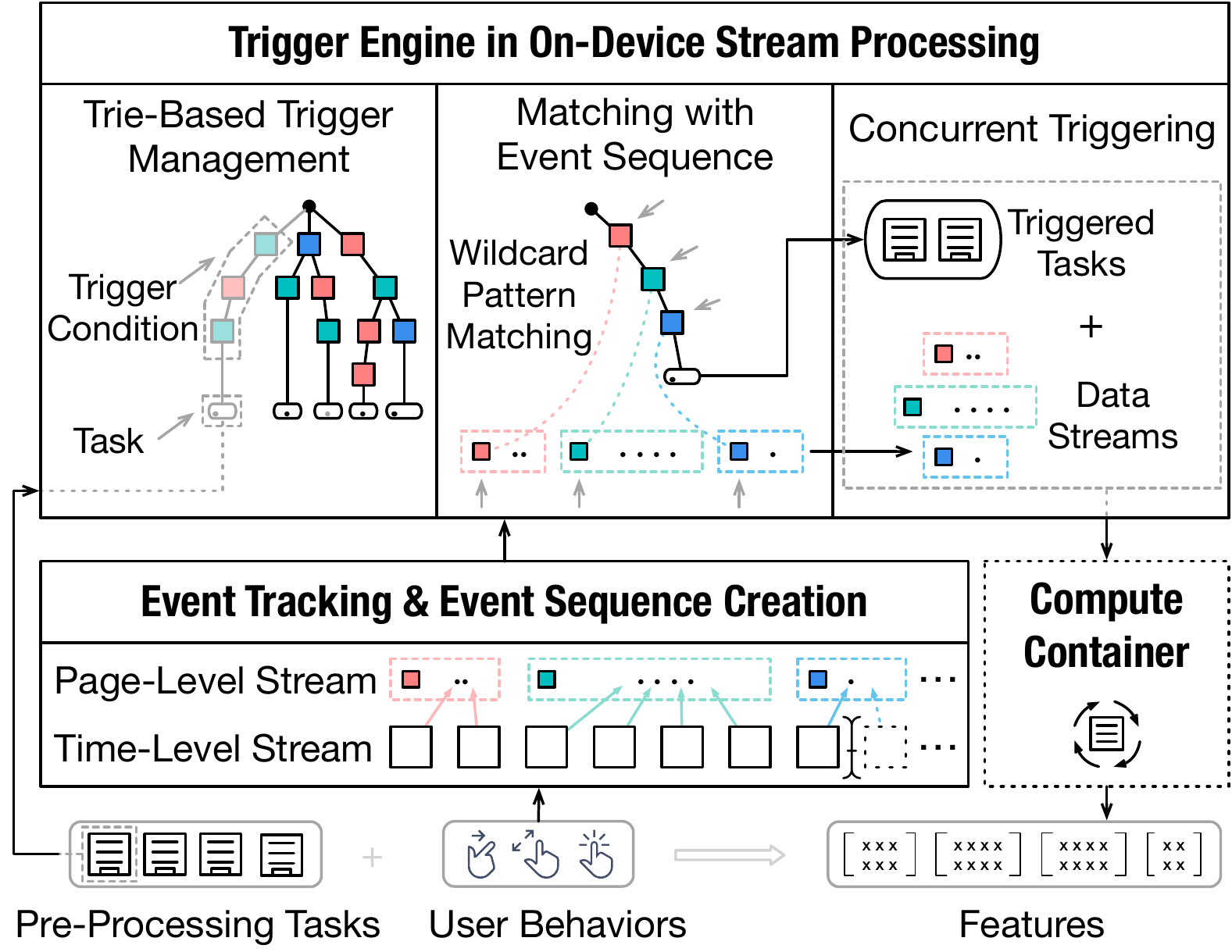}
    \vspace{-0.1em}
    \caption{Workflow of on-device stream processing.}
    \label{fig:device_stream_process_diagram}
    \vspace{-1.5em}
\end{figure}

\vspace{-0.4em}
\subsection{On-Device Stream Processing Framework}

The key design goal is to support stateful computation over unbounded data stream on single device. A user's behavior data in a mobile APP tracked with accurate timestamps form stream. The processing of user behavior data is stateful, where the intermediate results are buffered in memory or stored locally for later usage. The resources of single device are limited, which implies that the trigger conditions of many stream processing tasks should be well managed. We introduce on-device stream processing from event sequence creation, trigger management, task triggering, task execution, and collective storage. The workflow is depicted in Figure \ref{fig:device_stream_process_diagram}. 

{\bf Event Sequence Creation.} When a user interacts with a mobile APP, the user's behaviors will be tracked as events. There are five major kinds of basic events: page enter, page scroll, exposure, click, and page exit. Each kind of event is recorded with a unique event id, a page id, a timestamp, and event contents (\eg, the item id for exposure-type event and the graphical widget id for click-type event). Since a user's behaviors are naturally in time series, the time-level event sequence can be directly created. To further benefit processing the events within a certain page or cross pages, the page-level event sequence is created by aggregating the events between the enter and exit events of the same pages.

{\bf Trigger Management.} A stream processing task over the event sequence contains scripts and configurations, where the scripts implement the data processing algorithm, and the configurations mainly include a trigger condition. In particular, the trigger condition can be specified by a sequence of trigger ids, where a trigger id can be an event id or a page id. 

For a certain mobile device, it needs to efficiently maintain multiple pre-processing tasks to generate different features for diverse ML tasks, such that as an event comes, all relevant tasks can be triggered immediately. The key is to organize trigger conditions for quick matching. The trivial method of storing trigger conditions in a list is inefficient, because of the need to traverse the entire list each time. In fact, the matching of multiple trigger id sequences with the event sequence (with both event and page ids) can be modeled as a string matching problem with multiple wildcard patterns. Therefore, we leverage the data structure of prefix tree, called a trie, for efficient trigger management. More specifically, the trie has three kinds of nodes: start, middle, and end nodes. The trie's root is the unique start node. A trigger id is a middle node. An end node, which stores the stream processing tasks, is a leaf node of the trie, and vice versa. When a new stream processing task comes, the trigger id sequence will be extracted as a sequence of middle nodes, and a pair of start and end nodes will be added to the first and the last places of the node sequence, respectively. Then, the depth-first search is performed over the current trie from the root. If a path is completely matched with the node sequence, then the stream processing task will be added to the leaf node; otherwise, the mismatched nodes will be added to the trie as a new sub-tree, the root of which is the last matched node in the depth-first search process. We note that each path of the trie corresponds to a unique trigger condition, and the leaf node stores the stream processing tasks with the same trigger condition. If two trigger id sequences have common prefixes, then they will be put in the same sub-tree, and the middle nodes in the path from the trie's root to the sub-tree's root correspond to the common trigger ids.

{\bf Task Triggering.} When a new event (with an event id and a page id) comes, the set of triggered tasks will be returned. First, two lists of trie nodes are introduced to record the concurrent matching states of multiple trigger conditions and to avoid being blocked by wildcard pattern matching. The static pending list stores all the children of the trie's root, which correspond to the first trigger ids in all the trigger conditions and always keep active for matching. The dynamic pending list stores the desired next nodes of the trigger conditions in the ongoing matching. For an event in the stream, if its event/page id matches the trigger id of any node in the static or dynamic list, then each child of the node will be checked for whether it is an end node. If the child is an end node, then the stream processing tasks in the end node will be returned; otherwise, the child, as a new desired next node, will be added to a buffer of the dynamic list. At the end of task triggering for the event, the dynamic list will be replaced by the buffer, and the buffer will be refreshed. 

{\bf Task Execution.} When a task is triggered, the scripts will be run in the compute container to process relevant events. Besides standard data processing and mode execution APIs of the compute container, to facilitate the extraction of relevant events from the event sequence and the processing of event contents, the stream processing framework also provides some basic functions as follows: (1) {\em KeyBy}, which returns the events matched with a given key; (2) {\em TimeWindow}, which returns the events in a given time window; (3) {\em Filter}, which returns the events filtered by a defined rule; and (4) {\em Map}, which processes the event contents with a defined function.  

{\bf Collective Storage.} For each stream processing task, its outputs, typically features, are saved as a table using SQLite. Considering the fact that a stream processing task can be triggered for several times, while the size of one-time output is small, a collective data storage API is encapsulated over SQLite to reduce the number of write, thereby improving performance. In particular, a buffering table will be created in memory, and the output of a stream processing task is first written to the buffering table. If the number of write reaches a certain threshold or a read operation is invoked, the buffering table will be written into the database once.

\subsection{Real-Time Device-Cloud Tunnel}

Besides for local use, the output of on-device stream processing can also be uploaded to the cloud for real-time use. We implement a device-cloud tunnel based on the persistent connection. The secure sockets layer (SSL) protocol is optimized to reduce the time of connection establishment, encryption, and decryption. The data are compressed before transfer and are decompressed after transfer. To deal with high throughput, a fully asynchronous service framework is built on the cloud.

\begin{figure}[!t]
    \centering
    \includegraphics[width=.96\columnwidth]{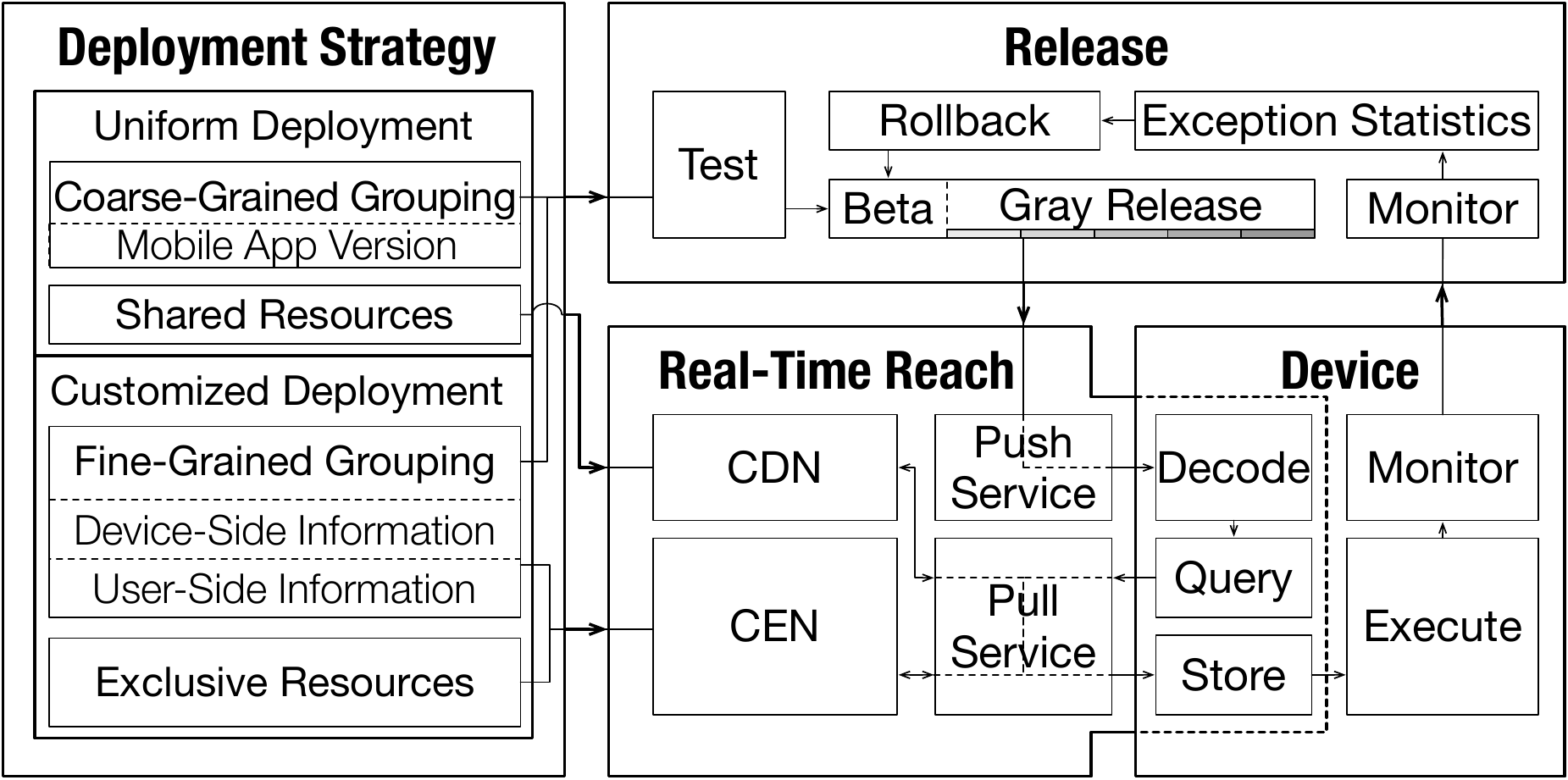}
    \vspace{-0.2em}
    \caption{Workflow of deployment platform.}\label{fig:deploy}
    \vspace{-1.1em}
\end{figure}

\vspace{-0.4em}
\section{Deployment Platform}\label{sec:walle:deployment:platform}
\vspace{-0.2em}

We introduce the details of ML task management, release, and deployment. The whole workflow is shown in Figure~\ref{fig:deploy}.

{\bf Task Management.} Git \cite{link:git} is adopted to achieve the isolation of different tasks and the version control of a certain task, while supporting collaborative development with access control. In particular, the entire task management is regarded as a git group; each business scenario corresponds to a git repo (repository); each task in a business scenario corresponds to a branch; and each version of a task corresponds to a tag.

Besides the management of task entity, the task-related files, especially the resources (\eg, data and model) which can be large in size, are also managed in a fine-grained way to support uniform and customized deployments. The files are divided into two categories: one is the shared files, which can be used by a large number of mobile devices (\eg, the devices with a certain version of APP); and the other is the exclusive files, which can be used only by a small number of devices or even a specific device. The shared and exclusive files can be requested efficiently via CDN and CEN, respectively. 

{\bf Task Release \& Deployment.} A uniform or customized policy can be taken to deploy tasks on targeted devices. The uniform policy supports task release grouped by the APP version, while the customized policy can further support grouping by device-side information (\eg, OS and its version or device performance) and user-side information (\eg, age or habit). According to the number of devices in a group, the coarse-grained uniform deployment normally involves only shared files, whereas the fine-grained customized task deployment not only can involve shared files but also can involve exclusive files. In the extremely personalized scenarios, the customized policy supports deploying a certain kind of task but with user-specific/exclusive files to each individual device.

Regarding task release, we take a novel push-then-pull method. We reuse the existing client-side business request to implement push, by adding a mobile device's local task profile into the http header and letting the cloud compare it with the latest task profile. If a new task needs to be released and takes the uniform deployment policy, then the cloud responds with the CDN address of the shared task files. If the new task takes the customized deployment policy, the cloud first determines which group the mobile device belongs to by rule matching and then responds with the CDN address of the shared files or the CEN address of the exclusive files. After receiving the response from the cloud, the mobile device can pull the task files using the CDN or CEN address from the nearest node. Considering the fact that the client-side business request is frequent, while the speeds of CDN and CEN are fast in practice, the timeliness of task deployment can be guaranteed. 

To guarantee the stability of task release and deployment, the simulators of the mobile APP with different versions for different OS can be created with the compute container on the cloud for testing a pre-release task extensively. Upon passing the simulation testing, a beta release is conducted to deploy the task only on a few targeted devices. After passing the beta release, the gray release is forced to be performed in steps, covering all the targeted devices incrementally. The deployment platform is also equipped with an exception handling module, which can monitor the failure rate of the task in real time and also can rollback immediately if the failure rate exceeds a certain threshold.

%-------------------------------------------------------------------------------
\vspace{-0.3em}
\section{Evaluation of Walle}
\vspace{-0.2em}
%-------------------------------------------------------------------------------

We evaluate Walle in two major application scenarios of \alibaba. We also extensively conduct benchmark testing for MNN, Python thread-level VM, and real-time tunnel. We finally report the statistics of the deployment platform.

\vspace{-0.4em}
\subsection{Performance in E-Commerce Scenarios}
\vspace{-0.1em}

{\bf Compute Container in Livestreaming.} E-commerce live-streaming has brought a brand new form of online shopping to billion-scale users. In 2020, the gross merchandise value (GMV) of livestreaming in \taobao\ exceeded 400 billion RMB. One key ML task in this scenario is highlight recognition, which is to locate the time points of a streamer in introducing attractive information about items.

Under the conventional cloud-based ML paradigm, a video stream is uploaded from each streamer's mobile device to the cloud for highlight recognition, which mainly includes the detection and recognition of items as well as the facial detection and the voice detection of streamers. Due to the large number of online streamers, the long length of their video streams, and the stringent latency requirement of highlight recognition, the load of the cloud is so heavy that only part of video streams and only a few sampled frames can be analyzed, which becomes a key bottleneck in practice.

\begin{figure}[!t]
    \centering
    \includegraphics[width=.9\columnwidth]{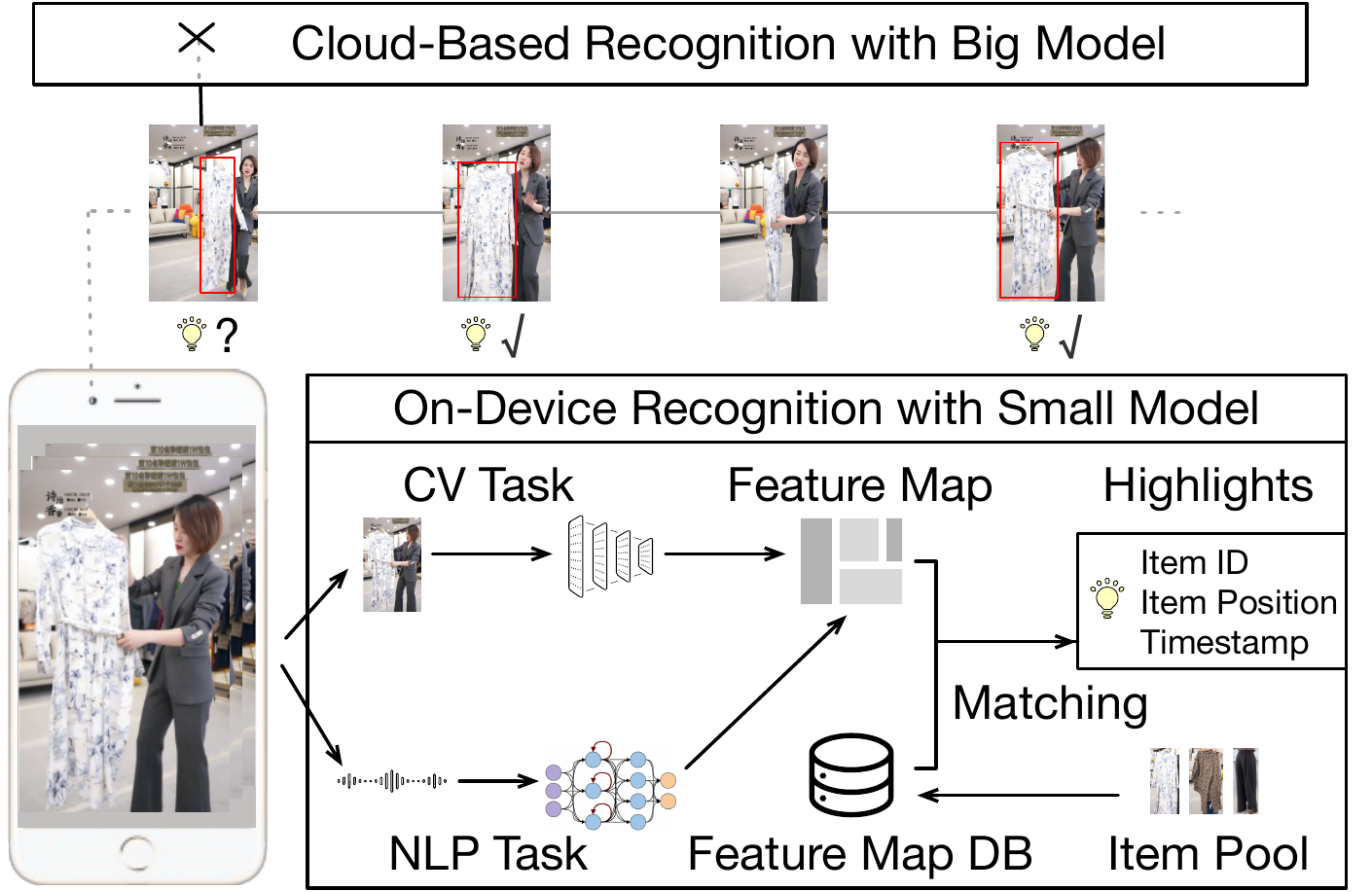}
    \vspace{-0.1em}
    \caption{Workflow of device-cloud collaborative highlight recognition in e-commerce livestreaming.}
    \vspace{-0.2em}
    \label{fig:highlight_recognition}
\end{figure}

\begin{table}[!t]
    \centering
    \caption{Model information and inference latency in device-side highlight recognition.}
    \label{tab:highlight}
    \resizebox{\columnwidth}{!}{
    \begin{tabular}[t]{l c c c c}
        \toprule
          & \tabincell{c}{Item\\ Detection} &
          \tabincell{c}{Item\\ Recognition} &
          \tabincell{c}{Facial\\ Detection} & \tabincell{c}{Voice\\ Detection}\\
        \midrule
        Model & FCOS \cite{proc:iccv19:fcos} & MobileNet \cite{arxiv17:mobilenet} & MobileNet \cite{arxiv17:mobilenet} & RNN\\
        Parameter Size & 8.15M & 10.87M & 2.06M & 8K\\
        \midrule
        Huawei P50 Pro & 56.92ms & 25.68ms & 41.42ms & 0.07ms \\
        iPhone 11 & 33.71ms & 29.74ms & 22.58ms & 0.01ms \\ 
        \bottomrule
    \end{tabular}
    }\vspace{-1em}
\end{table}

\begin{figure*}[!t]
    \centering
    \includegraphics[width=1.26\columnwidth]{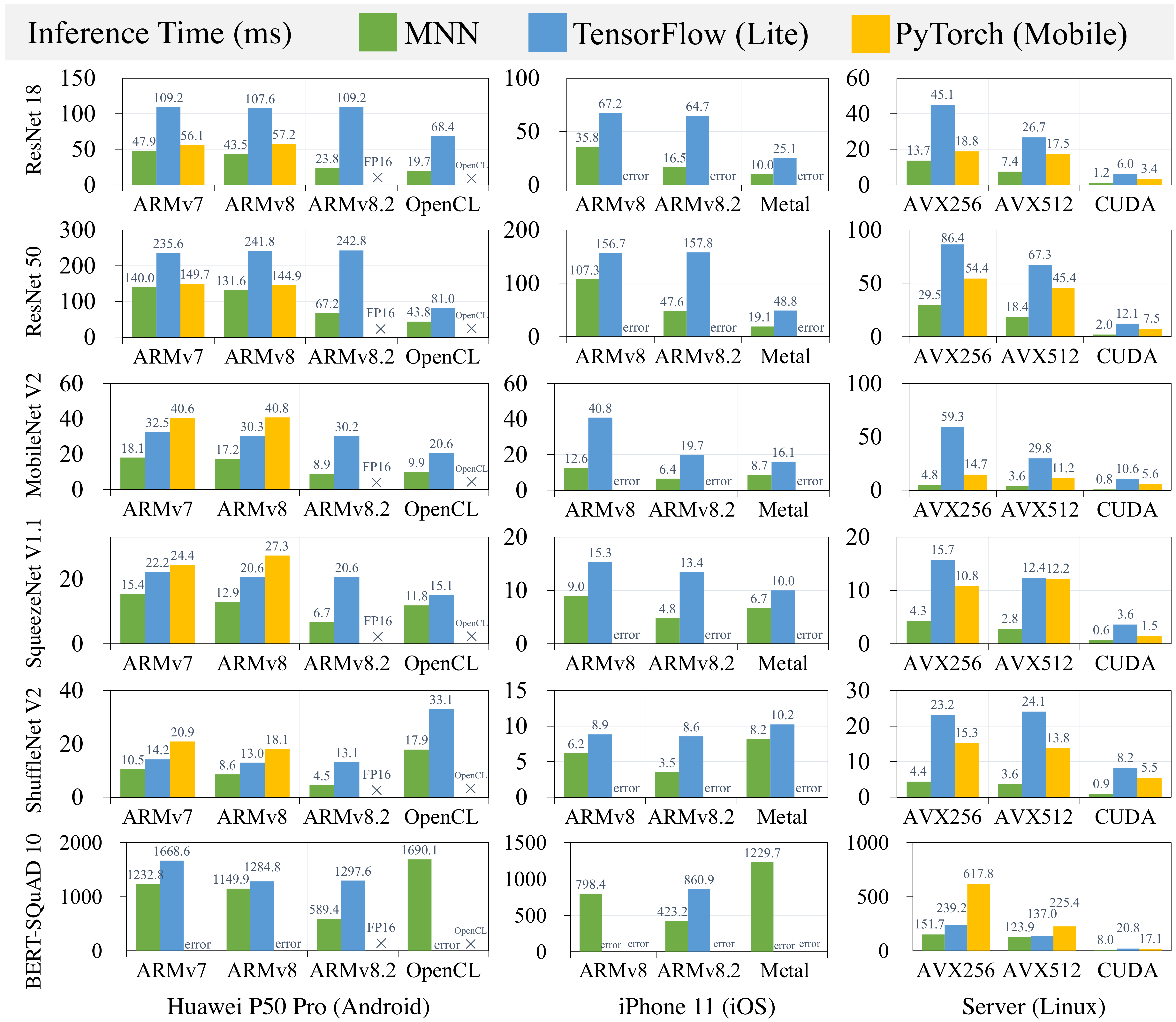}
    \includegraphics[width=.82\columnwidth]{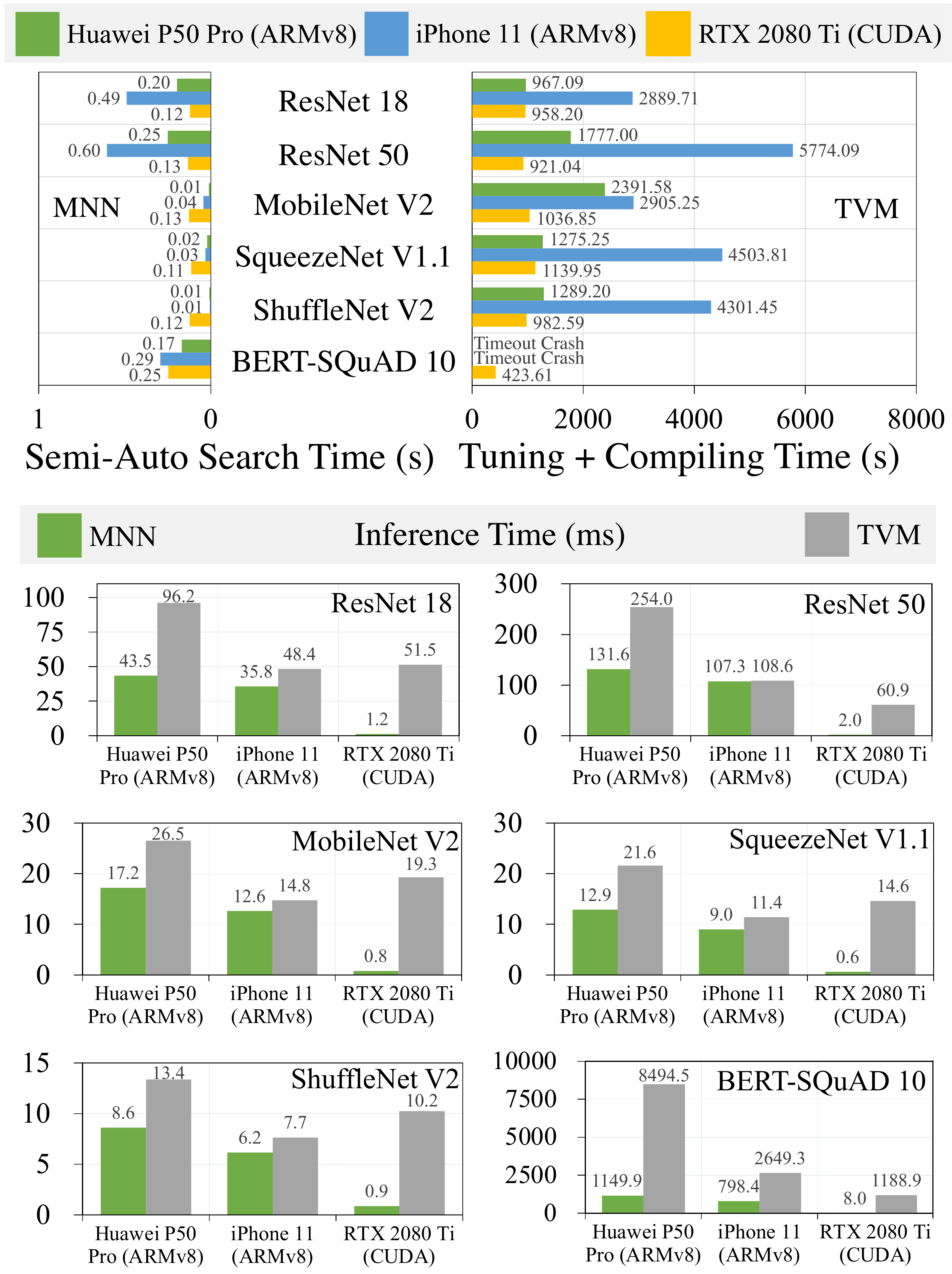}
    \vspace{-1.6em}
    \caption{MNN vs. TensorFlow (Lite), PyTorch (Mobile), and TVM on different backends of mobile devices and cloud servers.}\label{fig:mnn_benchmark}
    \vspace{-1.2em}
\end{figure*}

With Walle, we can offload the highlight recognition task with light-weight models to a streamer's mobile device and implement a device-cloud collaborative workflow, as shown in Figure~\ref{fig:highlight_recognition}. If the device-side models can recognize the highlights in a video stream with high confidences, then these highlights can be directly shown to users after post-processing. Only those highlights, which are recognized with low confidences on the mobile device and account for roughly 12\% in practice, need to be processed by cloud-side large models. After passing cloud-side recognition, the rate of which is around 15\%, the highlights will be delivered to the mobile device. 

Through device-cloud collaboration, the numbers of streamers and video streams covered with highlight recognition dramatically increase, while the load of the cloud is also sharply relived. In particular, business statistics show that compared with the cloud-based paradigm, the new device-cloud collaborative workflow increases the number of streamers with highlight recognition by 123\%; reduces the computing load of the cloud per highlight recognition by 87\%; and increases the size of daily recognized highlights per unit of cloud cost by 74\%. We also evaluate the performance of the compute container in Walle, when supporting highlight recognition on Huawei P50 Pro and iPhone 11. The total latency is 130.97ms and 90.42ms, respectively. In particular, the network architectures, the parameter sizes, and the inference latency of the adopted models are listed in Table \ref{tab:highlight}. The results above demonstrate the high performance of our compute container and the practical effectiveness of device-cloud collaboration.

{\bf Data Pipeline in Recommendation.} In \alibaba's cloud-side and device-side recommendation models, item page-view (IPV) feature, which records a user's behaviors (\eg, add-favorite, add-cart, and purchase) in the detailed page of an item, is of significant importance. To generate IPV feature, under the conventional cloud-based paradigm, all the users' raw event data are uploaded to the cloud for stream processing and mixed with user ids for explicit identification. The time-level event sequence from each mobile device is split into multiple homogeneous sequences, one sequence containing a certain kind of event. To obtain the IPV feature of each individual user, the cloud performs join operations with user id and page id as keys over all the users' events, which is quite time-consuming, resource-consuming, and error-prone.

With the on-device stream processing framework in the data pipeline, each mobile device needs to process only a small size of the corresponding user's local events, which is more efficient and more natural. In fact, the IPV feature invokes the generation process of the page-level event sequence. The input is the time-level event sequence. The trigger condition is the page exit event. The triggered stream processing task is to aggregate all the events (\ie, to cluster the same kind of events and gather statistics between the enter event and the exit event of the page). Since the raw contents in each event contain redundant fields (\eg, device status), a filtering is applied to the event contents. Further considering the fact that the IPV feature is first encoded (\eg, through RNN) in recommendation models, by using the model inference API of the compute container, the encoding process can also be offloaded to mobile devices.

We first show the size reductions from raw event data, to IPV feature, and to IPV encoding. On average, one IPV feature is around 1.3KB in size, involving 19.3 raw events in the size of 21.2KB, and one IPV encoding is only 128 bytes. This indicates that compared with the conventional paradigm of transferring raw event data to the cloud for stream processing, our new IPV data pipeline can save more than $90\%$ of communication cost. Besides communication efficiency, we also compare the latency of on-device and cloud-based stream processing. By analyzing over 10,000 practical cases (randomly sampled from the case pool of 2 million online mobile clients) of processing raw events into IPV features, the average on-device latency is only 44.16ms. In contrast, using Alibaba's internal version of Flink, called Blink, the average latency of producing one IPV feature is 33.73s. In particular, the cloud-based stream processing is over 2 million online users' raw events and consumes 253.25 compute units (CU), where 1 CU denotes 1 CPU Core plus 4GB memory; the error rate of IPV feature generation is 0.7\%; and the average latency is analyzed over 10,000 randomly sampled normal cases. These results reveal that compared the mainstream cloud-based data pipeline, Walle's new data pipeline can indeed reduce device-cloud communication cost and cloud-side load, while improving the timeliness and validity of feature.

\begin{figure*}[!t]
    \centering
    \begin{minipage}{0.33\textwidth}
        \centering
        \includegraphics[width=0.98\textwidth]{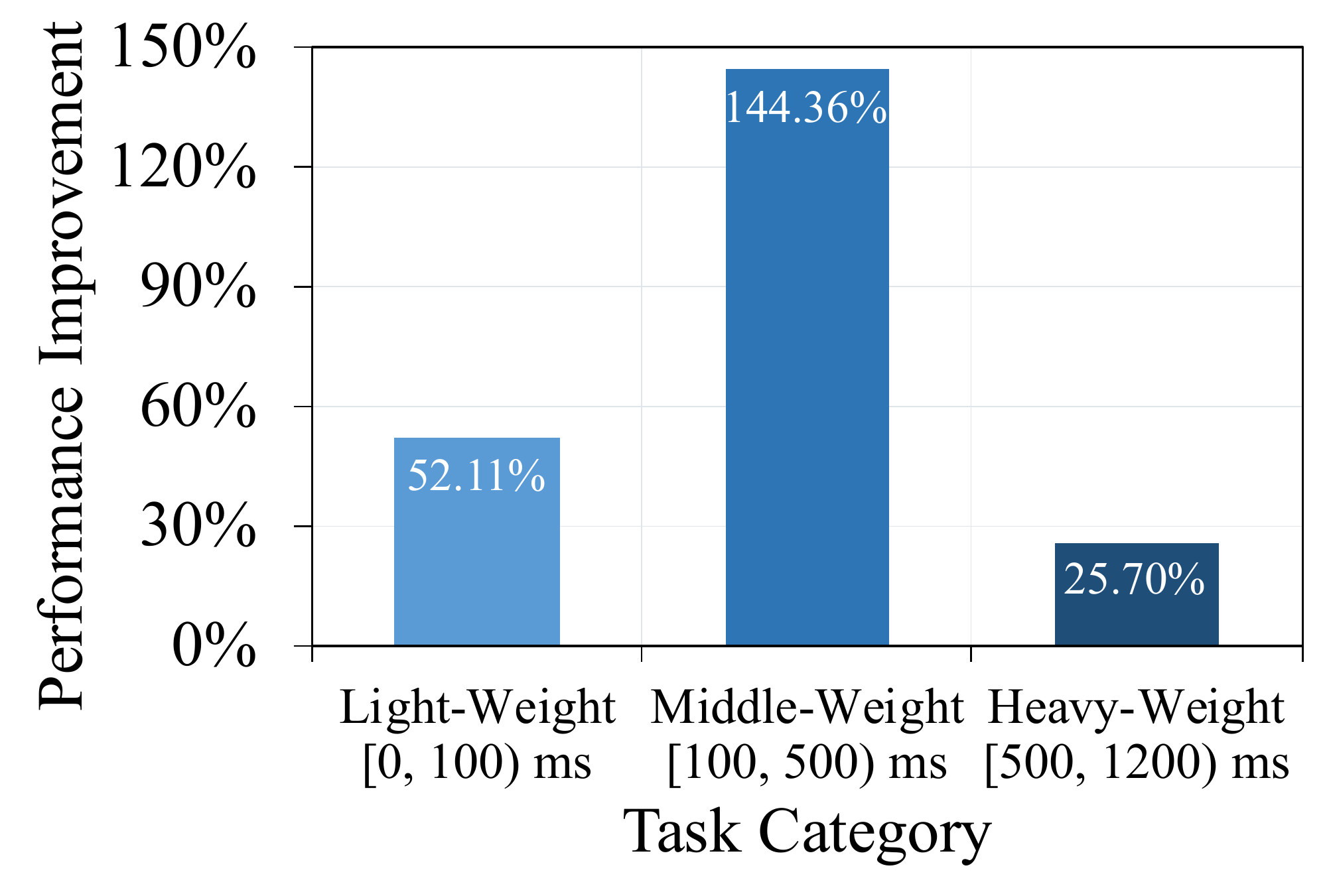}
        \vspace{-0.6em}
        \captionsetup{justification=centering}
        \caption{Python thread-level VM vs. \quad\ \ \\CPython (30 million ML task executions).}\label{fig:pythonvm_improvement}
        \vspace{-1.2em}
    \end{minipage}%
    \hfill
    \begin{minipage}{0.35\textwidth}
        \centering
        \includegraphics[width=1.0\textwidth]{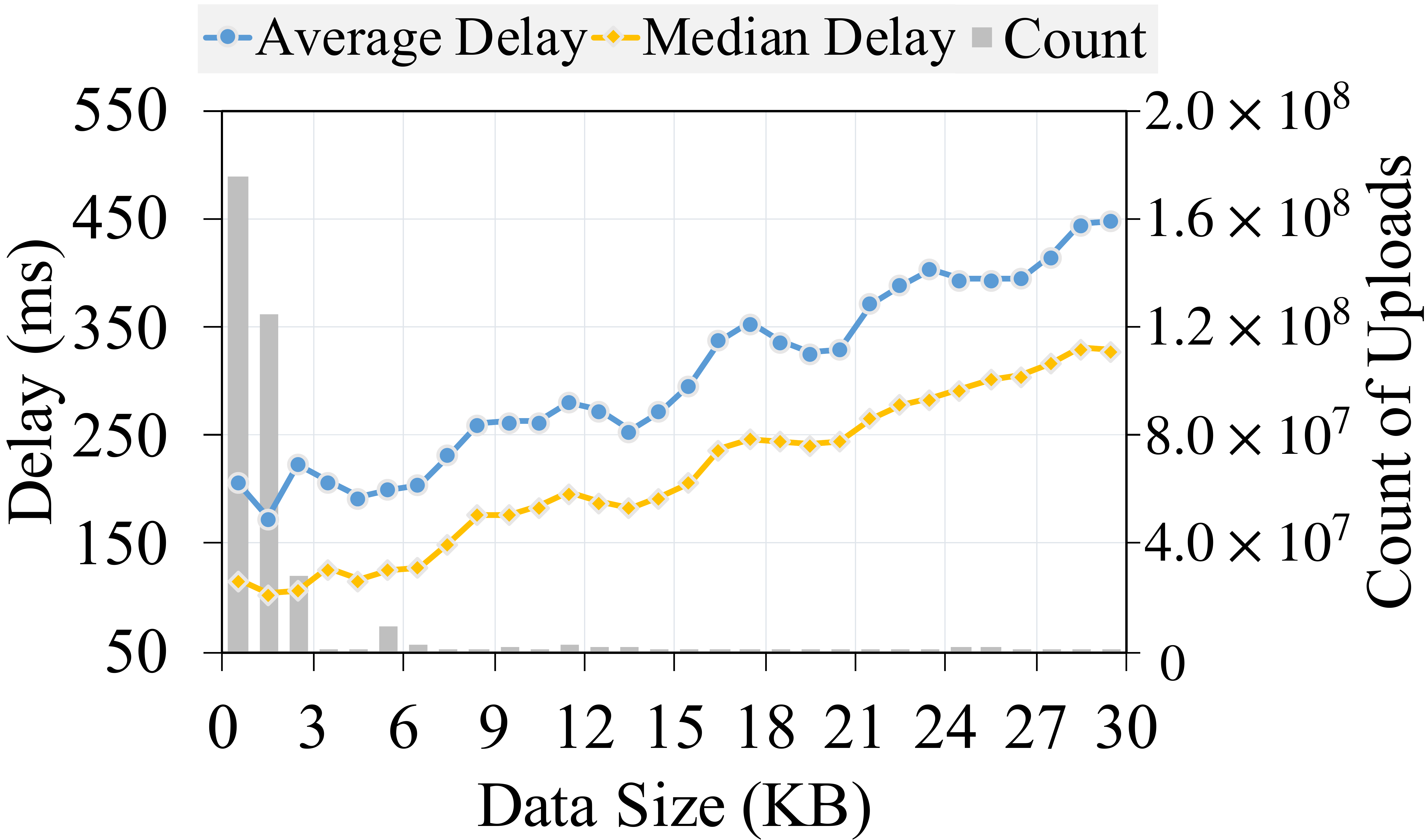}
        \vspace{-1.3em}
        \captionsetup{justification=centering}
        \caption{Delay of real-time tunnel \\ (analyzed over 364 million uploads).}\label{fig:real-time_tunnel}
        \vspace{-1.2em}
    \end{minipage}%
    \hfill
    \begin{minipage}{0.3\textwidth}
        \centering
        \includegraphics[width=1.0\textwidth]{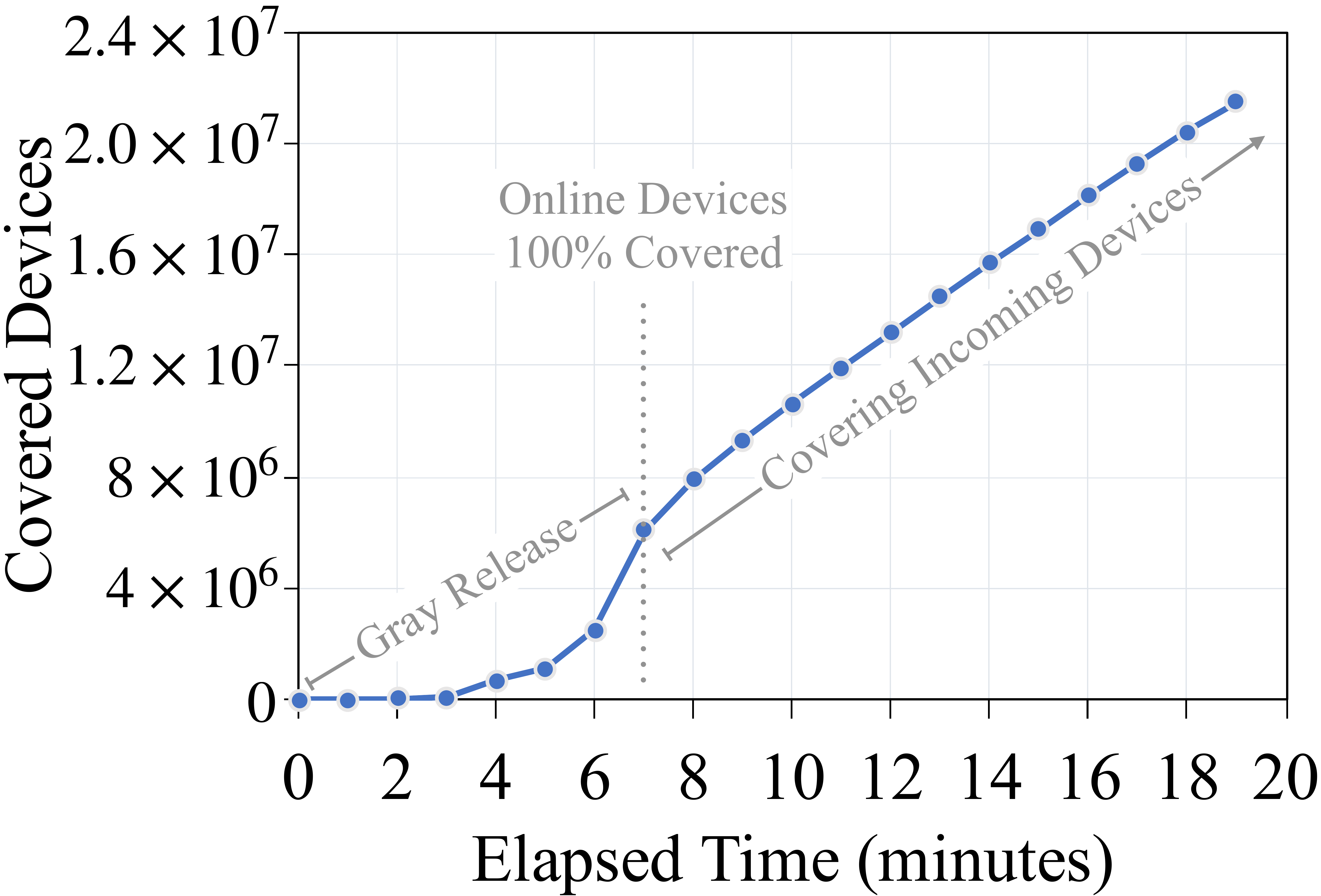}
        \vspace{-1.3em}
        \captionsetup{justification=centering}
        \caption{Timeliness of ML task deployment (22 million devices).}\label{fig:deployment_efficiency}
        \vspace{-1.2em}
    \end{minipage}
\end{figure*}

\vspace{-0.6em}
\subsection{Benchmark Testing}
\vspace{-0.3em}

We first compare MNN with TensorFlow (Lite) and PyTorch (Mobile) on Android and iOS devices as well as Linux servers. For device-side testing, we use Huawei P50 Pro and iPhone 11, covering the backends of ARMv7, ARMv8, and ARMv8.2 with single thread as well as OpenCL and Metal. For server-side testing, we use AMD Ryzen 9 3900X (x86), Alibaba Cloud's ecs.g6e.4xlarge (Intel Xeon (Cascade Lake) Platinum 8269CY, 16 vCPU, 64GiB memory), and NVIDIA GeForce RTX 2080 Ti, covering the backends of AVX256 and AVX512 with 4 threads, and CUDA, respectively. We take ResNet 18 \cite{proc:cvpr16:resnet}, ResNet 50 \cite{proc:cvpr16:resnet}, MobileNet V2 \cite{proc:cvpr18:mobilenetv2}, SqueezeNet V1.1 \cite{arxiv16:squeezenet}, ShuffleNet V2 \cite{proc:eccv18:shufflenetv2}, BERT-SQuAD 10 \cite{arxiv18:bert}, and DIN \cite{proc:kdd18:din}, as the testing models, which are commonly used in CV, NLP, and recommendation applications. The input size of CV models is set to $1 \times 3 \times 224 \times 224$, the input size of BERT-SQuAD 10 is set to $(1\times 256, 1\times 256, 1\times 256, 1)$, while the input size of DIN is set to $1 \times 100 \times 32$. We show the inference time of the CV and NLP models in the left part of Figure \ref{fig:mnn_benchmark}, and omit the results of DIN, which are quite low (\eg, less than 0.2ms on iPhone 11 using MNN). We can observe that MNN significantly outperforms TensorFlow (Lite) and PyTorch (Mobile) in almost all the test cases. Besides higher performance, MNN is also more full-featured on the side of mobile devices, given that MNN can support all the models on each device-side backend, whereas TensorFlow Lite and PyTorch Mobile fail to support some backends and/or models. 

We continue to compare MNN with TVM. We take MacBook Pro 2019 and NVIDIA GeForce RTX 2080 Ti as the host machines of TVM to do auto-tuning and compiling for the mobile devices and the GPU server, respectively. The number of trials in TVM auto-tuning is set to 30. Since TVM auto-tuning for BERT-SQuAD 10 on two mobile devices incurs timeout crash, we take the default parameter settings for model inference. From the evaluation results depicted in the right part of Figure \ref{fig:mnn_benchmark}, one key observation is that the auto-tuning and the compiling of TVM roughly cost thousands of seconds. In contrast, the semi-auto search of MNN for runtime optimization costs roughly hundreds of milliseconds. Further incorporating the comparative analysis in Section \ref{subsec:tensor:compute:engine}, we can draw that MNN can support the industrial scenarios that involve numerous heterogeneous devices and require frequent and quick task iteration, whereas TVM cannot. The second key observation is that the inference time of MNN is lower than TVM for each model on each backend, especially on the GPU server. Such superiority is mainly due to the manual operator-level, backend-level optimization in MNN.

We next compare Walle's Python thread-level VM with the original Python VM (\ie, CPython with GIL) using roughly 30 million online ML task executions. We define performance as the reciprocal of task execution time and show the average performance improvement in Figure~\ref{fig:pythonvm_improvement}. For the light-weight, middle-weight, and heavy-weight tasks, Python thread-level VM gains 52.11\%, 144.36\%, and 25.70\% of performance improvement, respectively. We can draw that task-level multi-threading without GIL is the key of performance boosting.

We finally evaluate the latency of the real-time tunnel over roughly 364 million uploads. Figure~\ref{fig:real-time_tunnel} shows the latency and the number of uploads for varying data sizes. We can observe that more than 90\% uploads are under 3KB with less than 250ms on average. Even when the sizes of 0.1\% uploads grow to 30KB, the average delay increases only to around 450ms.

\vspace{-0.5em}
\subsection{Deployment Platform Statistics}
\vspace{-0.1em}

The deployment platform in Walle has supported 30+ mobile APPs (\eg, Mobile Taobao, AliExpress, Xianyu, Youku, Cainiao Guoguo) in \alibaba\ since the end of 2017, running for roughly 1,500 days. It has deployed 1,000+ kinds of ML tasks in total, each with 7.2 versions on average. Currently, the deployment platform is maintaining and monitoring 348 kinds of active tasks on more than 0.3 billion mobile devices. To demonstrate the timeliness of task deployment, we randomly select an ML task, monitor its release process, and depict in Figure~\ref{fig:deployment_efficiency} how the number of covered devices changes as the elapsed time grows. The first segment of the curve shows the gray release stage, which takes 7 minutes to cover all the 6 million online devices. In particular, roughly 4 million devices are incrementally covered within the last minute. Then, the number of covered devices increases as more mobile devices become online. Until 19 minutes later, almost 22 million devices have been covered. The statistics show the scalability and timeliness of the deployment platform in Walle.

%-------------------------------------------------------------------------------
\vspace{-0.3em}
\section{Related Work}\label{sec:related:work:device:cloud:ML}
\vspace{-0.3em}
%-------------------------------------------------------------------------------

In this section, we briefly review some related work in both academia and industry.

{\bf Cloud-Based ML System.} Many companies have built their ML systems on the cloud, which are backed by their cloud computing platforms, such as Amazon Web Services, Microsoft Azure, Alibaba Cloud, and Google Cloud. The architecture is clear and comprises the standard modules of data storage (\eg, HBase \cite{link:hbase} and HDFS \cite{link:hadoop}), batch and stream processing (\eg, Storm \cite{proc:sigmod14:storm}, Spark~\cite{proc:hotcloud10:spark}, and Flink \cite{jour:ideb15:flink}), ML engines (\eg, TensorFlow \cite{proc:osdi16:tensorflow}, PyTorch \cite{proc:nips19:pytorch}, and MXNet \cite{arxiv15:mxnet}), virtualization and containerization (\eg, KVM \cite{link:kvm} and docker \cite{link:docker}), and elastic orchestration (\eg, Kubernetes \cite{link:k8s}).

{\bf On-Device ML System.} Some modules are open source with rapid development in terms of well balancing light weight, necessary functionality, and high performance, including on-device inference engines (\eg, TensorFlow Lite \cite{link:tf:lite}, PyTorch Mobile \cite{link:pytroch:mobile}, Core ML \cite{link:core:ml}, and NCNN \cite{link:ncnn}); and SQLite \cite{link:sqlite}, which is a small and self-contained SQL database engine. However, the whole architecture is still in the dark, and several core capabilities are absent, such as an on-device execution environment that supports quick development and concurrent execution of multiple ML tasks, and light-weight data processing and model training libraries for diverse CV, NLP, and recommendation tasks.

{\bf Device-Cloud Collaborative ML.} The concept can stretch back to edge/mobile computing, but focuses on the collaboration of the cloud and mobile devices in executing complex ML tasks, rather than offloading simple data analysis tasks from the cloud to edge servers or mobile devices. Previous work focused on the algorithmic framework or solution and was normally specific to a certain kind of application. In contrast and in parallel, Walle targets at the general-purpose and large-scale production system support. We review some representative work as follows. 

An initial paradigm is to keep model training on the cloud but offload model inference (\eg, facial recognition, photo beautification, and question answering) to mobile devices, validating the on-device advantages in reducing latency and protecting privacy. The key of this paradigm's proliferation is the advances of model compression algorithms to reduce model size and optimize model structure, such as quantization \cite{proc:icml15:quantization}, pruning and sparification \cite{proc:nips15:pruning}, knowledge distillation \cite{arxiv15:knowlege:distill}, and neural architecture search~\cite{proc:iclr17:nas}. Later, Mistify \cite{proc:nsdi21:mistify} automated the cloud-to-device model porting process given the customized requirements of heterogeneous mobile devices, while some work designed more reasonable task splitting strategies rather than offloading the full inference task. For example, Neurosurgeon \cite{proc:asplos17:dnn:split} was proposed to automatically partition DNN computation between a mobile device and the cloud at the granularity of DNN layers. 

Besides inference, the popular cross-device federated learning (FL) framework \cite{proc:aistats17:fl} elegantly generalizes the conventional parameter server framework \cite{proc:osdi14:li:ps} and enables multiple mobile devices to collaboratively train a global model under the coordination of a cloud server. The tenet of FL is to keep user data on local devices, thereby protecting data security and privacy. The device-cloud collaboration in FL is purely through exchanging model and its update periodically. The task splitting strategy is that mobile devices conduct model training, and the cloud aggregates model updates. Google has experimentally deployed FL on its Android keyboard, called Gboard, to polish language models \cite{proc:mlsys19:fl:system}. 

Finally, many application-specific solutions were proposed under the principle of device-cloud collaboration. FilterForward \cite{proc:mlsys19:video:edge:classifier} and Reducto \cite{proc:sigcomm20:video:diff} considered how to effectively and efficiently do camera-side frame filtering with ML techniques to facilitate cloud-side video analytics. DDS \cite{proc:sigcomm20:video:camera:resend} adopted an interactive workflow, where a camera first uploads a low-quality video stream and re-sends a few key regions with higher quality according to the cloud's feedback to improve inference accuracy. COLLA \cite{proc:sec19:colla:distill} studied the user behavior prediction task with RNN and leveraged knowledge distillation to mutually and continuously transfer the knowledge between the device-side small models and the cloud-side large model, thereby mitigating data heterogeneity and data drift over time. DDCL \cite{proc:kdd21:patch} and CoDA \cite{arxiv22:coda} focused on recommendation. DDCL relied on patch learning for on-device model personalization and adopted model distillation to integrate the patches from mobile devices into the cloud-side global model. CoDA, instead, was proposed to retrieve similar samples from the cloud's global pool to augment each mobile device's local dataset for training personalized recommendation models. Backed by Walle, CoDA was deployed in Mobile Taobao.

%-------------------------------------------------------------------------------
\vspace{-0.7em}
\section{Conclusion}
\vspace{-0.3em}
%-------------------------------------------------------------------------------
In this work, we have built the first end-to-end, general-purpose, and large-scale production system, called Walle, for device-cloud collaborative ML. Walle is oriented by the lifecycle of ML tasks and consists of a cross-platform, high-performance, and quickly iterative compute container; a more reasonable and efficient data pipeline; and a scalable, timely, and robust deployment platform. Evaluation of Walle in practical e-commerce scenarios and extensive micro-benchmarks have demonstrated the necessity of device-cloud collaboration and the superiority of each ingredient. Walle has been deployed in \alibaba\ for wide scale production use, serving billion-scale users with mobile devices every day.

%-------------------------------------------------------------------------------
\vspace{-0.7em}
\section*{Acknowledgments}
\vspace{-0.3em}
%-------------------------------------------------------------------------------

We sincerely thank our shepherd, Wenjun Hu, for her insightful and thorough guidance. We thank the anonymous OSDI reviewers for their constructive feedback. We thank Kai Liu, Hansong Liu, Hao Jiang, Zhijie Cao, and Yan Chen from Alibaba for their great support. This work was supported in part by National Key R\&D Program of China No. 2019YFB2102200, China NSF grant No. 62025204, 62072303, 61972252, 61972254, 61832005, and 62141220, and Alibaba Innovation Research (AIR) Program. The opinions, findings, conclusions, and recommendations expressed in this paper are those of the authors and do not necessarily reflect the views of the funding agencies or the government.

%-------------------------------------------------------------------------------

\balance
\bibliographystyle{plain}
\bibliography{ref}

%%%%%%%%%%%%%%%%%%%%%%%%%%%%%%%%%%%%%%%%%%%%%%%%%%%%%%%%%%%%%%%%%%%%%%%%%%%%%%%%
\end{document}